\documentclass[letterpaper,journal]{IEEEtran}

\newif\ifanonymous
\anonymousfalse

\ifdefined\pdfsuppresswarningpagegroup
\fi

\usepackage{amsmath,amsfonts}
\usepackage{amssymb}
\usepackage{mathtools}
\usepackage{algorithmic}
\usepackage{algorithm}
\usepackage{array}
\usepackage[caption=false,font=normalsize,labelfont=sf,textfont=sf]{subfig}
\usepackage{textcomp}
\usepackage{stfloats}
\usepackage{url}
\usepackage{verbatim}
\usepackage{graphicx}
\usepackage{cite}
\usepackage{microtype}
\usepackage{booktabs}
\usepackage{tabularx}
\usepackage{multirow}
\usepackage{makecell}
\usepackage[table]{xcolor}
\usepackage{colortbl}
\colorlet{blue}{black}
\usepackage[hidelinks]{hyperref}
\usepackage[capitalize,noabbrev]{cleveref}

\ifanonymous
  \hypersetup{
    pdftitle={Dynamics-Aware Meta-Imitation for Generalization to Unseen Robotic Manipulation},
    pdfauthor={Anonymous Authors}
  }
\else
  \hypersetup{
    pdftitle={Dynamics-Aware Meta-Imitation for Generalization to Unseen Robotic Manipulation},
    pdfauthor={Zhenduo Shang, Xiyao Liu, Bohan Li, Xudong Wang, Teng Ren, Lianqing Liu, and Zhi Han}
  }
\fi

\newcolumntype{Y}{>{\centering\arraybackslash}X}

\newenvironment{IEEEpractitioners}{%
  \renewcommand{\abstractname}{Note to Practitioners}%
  \begin{abstract}%
}{%
  \end{abstract}%
  \renewcommand{\abstractname}{Abstract}%
}

\newcounter{appendixsection}

\crefname{appendixsection}{Appendix}{Appendices}
\Crefname{appendixsection}{Appendix}{Appendices}

\makeatletter
\def\bstctlcite#1{\@bsphack
  \@for\@citeb:=#1\do{%
    \edef\@citeb{\expandafter\@firstofone\@citeb}%
    \if@filesw\immediate\write\@auxout{\string\citation{\@citeb}}\fi}%
  \@esphack}
\makeatother

\title{Dynamics-Aware Meta-Imitation for Generalization to Unseen Robotic Manipulation}

\ifanonymous
  \author{Anonymous Authors}
\else
  \author{Zhenduo Shang, Xiyao Liu, Bohan Li, Xudong Wang, Teng Ren, Lianqing Liu, and Zhi Han%
  \thanks{Zhenduo Shang, Xiyao Liu, Xudong Wang, Lianqing Liu, and Zhi Han are with the State Key Laboratory of Robotics and Intelligent Systems, Shenyang Institute of Automation, Chinese Academy of Sciences, Shenyang 110016, China.}%
  \thanks{Zhenduo Shang and Xudong Wang are also with the University of Chinese Academy of Sciences, Beijing 100049, China.}%
  \thanks{Bohan Li and Teng Ren are with Shenyang University of Technology.}%
  \thanks{Corresponding author: Xiyao Liu (e-mail: liuxiyao@sia.cn).}%
  }
\fi

\begin{document}

\bstctlcite{IEEEexample:BSTcontrol}

\maketitle

\begin{abstract}
Imitation \textcolor{blue}{learning enables robots} to \textcolor{blue}{acquire} skills from observations and \textcolor{blue}{expert} demonstrations, \textcolor{blue}{but remains constrained by} data scarcity and \textcolor{blue}{limited} generalization. \textcolor{blue}{Existing} methods focus on \textcolor{blue}{fixed task sets} and struggle \textcolor{blue}{to adapt} to unseen tasks. To bridge this gap, we propose the Dynamics-Aware Meta-Imitation (DAMI) framework. \textcolor{blue}{It combines a three-dimensional diffusion policy with model-agnostic} meta-learning to \textcolor{blue}{learn} a shared \textcolor{blue}{policy initialization} for \textcolor{blue}{few-shot} adaptation. \textcolor{blue}{The} Visual-Motor Trajectory (VMT) module \textcolor{blue}{jointly encodes visual features, expert actions, and temporal positions from a randomly selected complete reference demonstration into an action-aware} spatio-temporal task \textcolor{blue}{representation}. \textcolor{blue}{The} Unpaired Unified Task (U2T) \textcolor{blue}{module fuses demonstration and language representations with current observation and diffusion timestep information.} Task-Conditioned Feature Modulation (TCFM) \textcolor{blue}{uses the fused condition to modulate} low-level features \textcolor{blue}{in the diffusion policy. Demonstration and language conditioning remain active during execution and adaptation.} \textcolor{blue}{We evaluate} direct inference on seen tasks and adaptation to unseen tasks \textcolor{blue}{using Meta-World, RLBench, and real-world robotic manipulation experiments}. \textcolor{blue}{The real-world experiments also assess generalization to unseen object instances within the same task. The framework achieves higher average success rates in several evaluated comparisons, supporting adaptation from limited demonstrations.}
\end{abstract}

\begin{IEEEpractitioners}
Deploying robot manipulators in factories and service environments is often limited by the cost of collecting demonstrations for every new task. This work \textcolor{blue}{addresses} the practical need to reuse experience from previously learned tasks and adapt a robot to a new manipulation task \textcolor{blue}{using} a \textcolor{blue}{limited set} of demonstrations. The proposed DAMI framework combines the robot's current observation, a language instruction, and a complete reference demonstration to \textcolor{blue}{guide action generation} and \textcolor{blue}{support task-specific adaptation}. Tests in simulated benchmarks and on a physical robot show \textcolor{blue}{higher} task success \textcolor{blue}{rates in several comparisons} with representative baseline methods. The method is most applicable when a complete and relevant reference demonstration, reliable three-dimensional observations, and a small task-specific demonstration set are available. Its present limitations include fine-tuning \textcolor{blue}{on unseen tasks}, additional \textcolor{blue}{computational requirements}, and evaluation on a limited set of manipulation tasks. Future work should reduce adaptation time and validate reliability under broader object, sensing, and operating conditions.
\end{IEEEpractitioners}


\begin{figure}[t]
	\centering
	\includegraphics[width=\columnwidth]{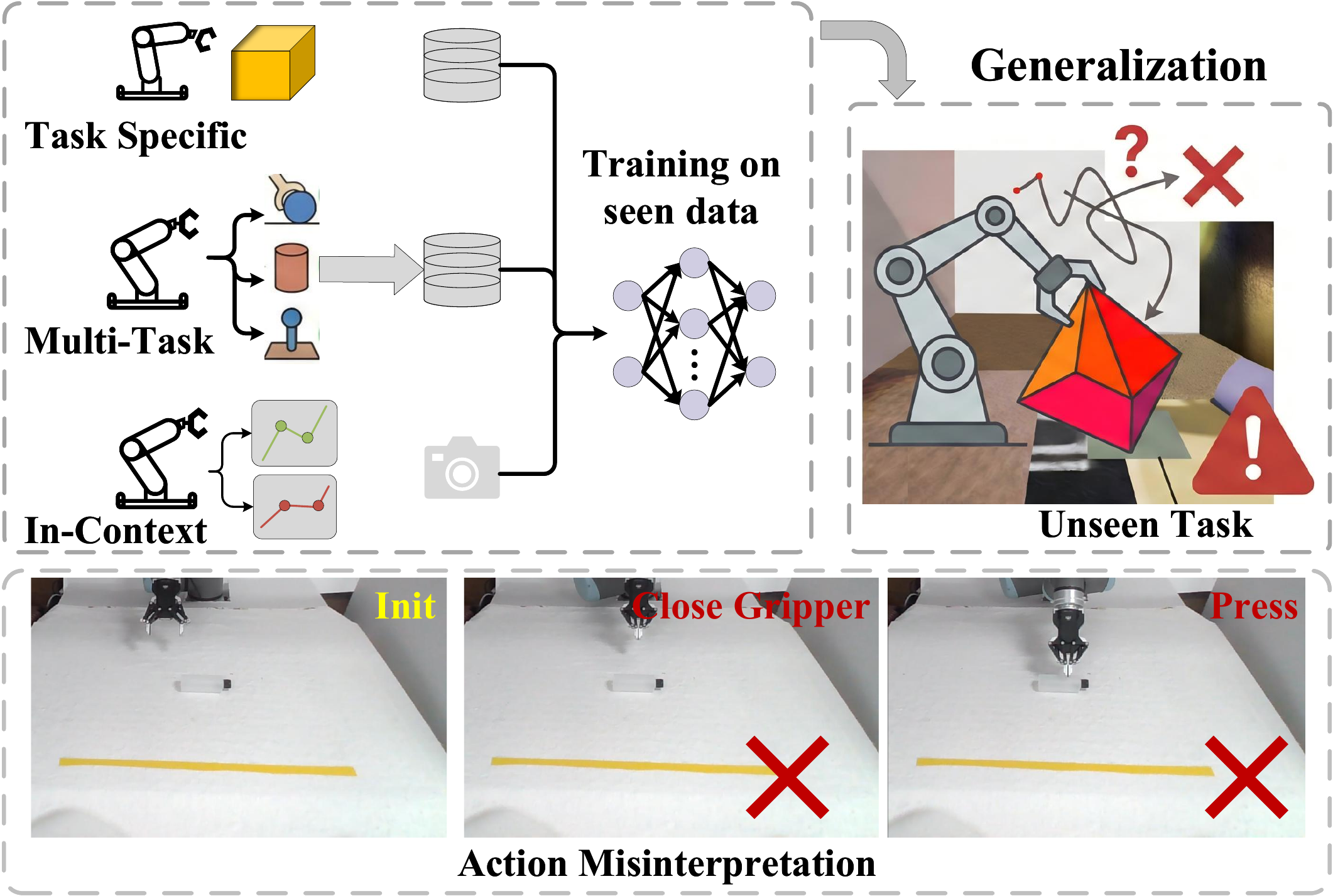}
	\caption{\textbf{\textcolor{blue}{Motivation for task-conditioned adaptation}.} \textcolor{blue}{Top: conceptual transfer from} seen \textcolor{blue}{to} unseen tasks. \textcolor{blue}{Bottom}: \textcolor{blue}{an action-misinterpretation example showing pressing instead of pushing} the cylinder.}
	\label{fig:motivation}
	\vspace{-0.4cm}
\end{figure}

\section{Introduction}

Robotic manipulation aims to enable agents to interact with and change their environments \cite{ji2025robobrain, li2025robotic}. Imitation Learning (IL) is effective for acquiring complex robotic manipulation skills from demonstrations \cite{lin2025adaptive, xu2025bikc+}. However, IL policies learned from limited expert demonstrations can suffer from distribution shift and compounding errors, while collecting additional demonstrations is often costly \cite{lin2026toward, wang2025robot, jiang2025dexmimicgen}. \textcolor{blue}{Existing approaches include specialized policies for individual tasks \cite{xia2025cage, su2026freqpolicy} and multi-task models trained across a fixed set of tasks \cite{goyal2023rvt, gervet2023act3d}. Generalization beyond the training tasks remains challenging, limiting their applicability to general-purpose robots \cite{team2024octo}.}

Large-scale human videos \cite{chen2025vividex} and synthesized data \cite{jang2025dreamgen} can reduce the need for robot demonstrations, but training \textcolor{blue}{costs} and domain shifts \textcolor{blue}{remain \cite{fang2025airexo}. Few-shot adaptation uses prior knowledge and a small target-task demonstration set to update a policy for a new task \cite{liu2025one, wang2025one}. Adaptation remains challenging when the target task differs substantially from the meta-training tasks.} In-Context Imitation Learning (ICIL) uses demonstrations directly as \textcolor{blue}{inference context without gradient updates \cite{oh2025robust, icilgd_iclr25}. These approaches motivate combining demonstration-guided policy execution with parameter adaptation to new tasks.}

\textcolor{blue}{Meanwhile,} 3D policy learning \textcolor{blue}{uses} explicit representations, including point clouds and fused geometric features, \textcolor{blue}{to capture} spatial structure in unstructured environments \cite{zhao2025robot, zhong2025region, yu2025fast}. Within this area, diffusion policies \textcolor{blue}{model} multimodal action distributions and \textcolor{blue}{generate} complex visuomotor behaviors \cite{chi2025diffusion, Ze2024DP3, ze2025generalizable}. However, fine-tuning a high-capacity 3D diffusion policy \textcolor{blue}{on limited target-task demonstrations can overfit the available trajectories \cite{chen2025adp, zhao2026robot}. Similar scenes can support different manipulation goals, so geometry alone may not specify the intended interaction. A complete reference demonstration supplies observations and expert actions over time to help resolve this ambiguity. \Cref{fig:motivation} illustrates this motivation through a task-transfer schematic and an action-misinterpretation example.}

\begin{figure*}
	\centering
	\includegraphics[width=\textwidth]{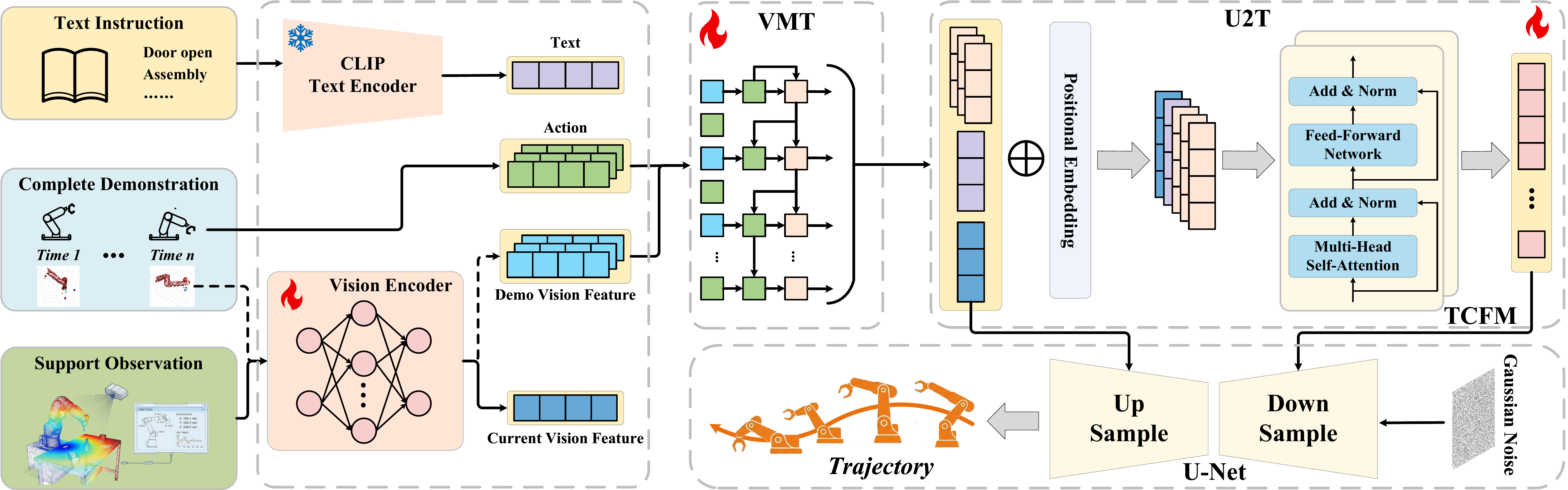}
	
	\caption{\textbf{Overview of the DAMI Architecture.} \textcolor{blue}{The policy uses} current \textcolor{blue}{observations}, language, and a complete reference demonstration\textcolor{blue}{. ``Support Observation'' denotes the current-observation branch used in support}, \textcolor{blue}{query, and execution forward passes}. \textcolor{blue}{Sections III-C--III-E describe the conditioning path, and Section IV-A provides implementation details. TCFM operates in both first-stage residual blocks.}}

	\label{fig:model}
	
	\vspace*{-0.4cm}
\end{figure*}

To this end, we present the Dynamics-Aware Meta-Imitation (DAMI) framework\textcolor{blue}{. It combines a 3D diffusion policy \cite{Ze2024DP3} with MAML \cite{finn2017model} to learn a shared policy initialization for novel-task adaptation. The initialization reuses experience across training tasks, while language and a complete reference demonstration provide task-specific conditions during execution and adaptation. The} Visual-Motor Trajectory (VMT) module jointly encodes visual features\textcolor{blue}{, expert actions, and temporal positions into an action-aware spatio-temporal representation of the reference} demonstration. The Unpaired Unified Task (U2T) \textcolor{blue}{module fuses this representation with language} and observation--timestep \textcolor{blue}{information, without requiring frame-wise alignment with current observations}. Task-Conditioned Feature Modulation (TCFM) \textcolor{blue}{uses the fused condition to modulate low-level features in} the diffusion U-Net \cite{ronneberger2015u, perez2018film}. \textcolor{blue}{This design combines demonstration-guided execution with few-shot parameter adaptation.} In summary, our contributions are as follows:

\begin{itemize}
	\item We \textcolor{blue}{develop} DAMI\textcolor{blue}{, a demonstration-conditioned 3D diffusion-policy} framework \textcolor{blue}{trained with MAML} for \textcolor{blue}{direct inference on base tasks and} few-shot adaptation \textcolor{blue}{to novel tasks}.

	\item We introduce VMT to \textcolor{blue}{jointly encode visual features, expert actions, and temporal positions from complete demonstrations into an action-aware} spatio-temporal task \textcolor{blue}{representation}.

	\item We design \textcolor{blue}{a} U2T\textcolor{blue}{--TCFM conditioning architecture that fuses demonstration and language representations} with \textcolor{blue}{observation--timestep information} to modulate low-level features \textcolor{blue}{in the diffusion U-Net}.

	\item Experiments on Meta-World, RLBench, and real-world tasks \textcolor{blue}{evaluate base-task} direct inference and \textcolor{blue}{novel-task performance after few-shot} adaptation.
\end{itemize}

\section{Related Work}


\subsection{Few-Shot Imitation Learning}

\textcolor{blue}{Visual meta-imitation~\cite{finn2017visualmetaimitation} uses gradient-based meta-learning~\cite{finn2017model} for adaptation from visual demonstrations, while recent work incorporates mixture-of-experts models~\cite{zhao2026robot}. Instant Policy~\cite{icilgd_iclr25} uses graph diffusion, while ICRT~\cite{fu2025icrt} uses causal next-token prediction over sensorimotor sequences. Both use demonstrations as inference context without target-task gradient updates. Other transfer mechanisms include multi-task learning in BC-Z~\cite{jang2022bc}, retrieval in RAM~\cite{kuang2024ram}, and structured constraints in OmniManip~\cite{pan2025omnimanip} and Funcanon~\cite{xu2025funcanon}. One-shot manipulation methods transfer strategies through contact analogies~\cite{liu2025one} or represent object relations and constraints with symbolic abstraction graphs~\cite{wang2025one}. IMOP~\cite{zhang2024imop} matches invariant regions across demonstrations and test scenes to transfer end-effector poses without new-task fine-tuning. RHyME~\cite{kedia2025one} addresses mismatches between demonstrated and robot executions, whereas CoVA-IL~\cite{luo2026cova} uses object-centric viewpoint alignment for zero-shot imitation. AutoLfD~\cite{wu2025autolfd} closes the learning loop by evaluating transfer quality and optimizing learning-from-demonstration hyperparameters. DAMI combines gradient-based task adaptation with continued conditioning on a complete reference demonstration during execution, connecting meta-imitation with demonstration-conditioned policies.}

\subsection{3D Visual-Motor Policies}
3D representations provide geometry and viewpoint invariance, improving policy robustness in robotic manipulation \cite{qi2017pointnet++}. DP3 \cite{Ze2024DP3} established a strong diffusion baseline \textcolor{blue}{for 3D robotic manipulation. Subsequent advances include} visual pre-training \cite{hou20254d}, feature fusion \cite{liu2025spatial}, joint viewpoint planning and depth completion \cite{liu2025rasp}, \textcolor{blue}{and multi-view graph attention for grasping \cite{ding2025dual}. Task-driven grasping also} uses part-affordance grounding and multimodal \textcolor{blue}{language-model reasoning \cite{song2025learning, zhao2026affordance}. ControlVLA~\cite{li2025controlvla} introduces object-centric conditions into a pretrained diffusion Transformer through ControlNet-style projections and fine-tuning. DemoDiffusion~\cite{park2026demodiffusion} refines retargeted human-demonstration trajectories with a pretrained diffusion policy, without task-specific training. DAMI uses a temporally encoded visual--action demonstration together with language and current observations to provide task conditions for a 3D diffusion policy.}

\subsection{Multi-task Robotic Manipulation}

RVT-2 \cite{goyal2024rvt} improves \textcolor{blue}{manipulation} precision with multi-view \textcolor{blue}{Transformers}, \textcolor{blue}{while} GNFactor \cite{ze2023gnfactor} uses neural feature fields for semantic understanding. ManiGaussian \cite{lu2024manigaussian} models scene dynamics with dynamic Gaussian Splatting \cite{kerbl20233d}, while diffusion-based frameworks \cite{ke20243d, tian2025pdfactor} and FlowRAM \cite{wang2025flowram} improve trajectory modeling and efficiency. \textcolor{blue}{MUTEX~\cite{shah2023mutex} learns a unified policy from video}, \textcolor{blue}{image, text,} and \textcolor{blue}{speech specifications through masked modeling and cross-modal matching}. \textcolor{blue}{RDT-1B~\cite{liu2025rdt1b} scales multimodal diffusion Transformers for bimanual manipulation and reports learning new skills from limited demonstrations.}

\begin{figure}[t]
	\centering
	\includegraphics[width=\columnwidth]{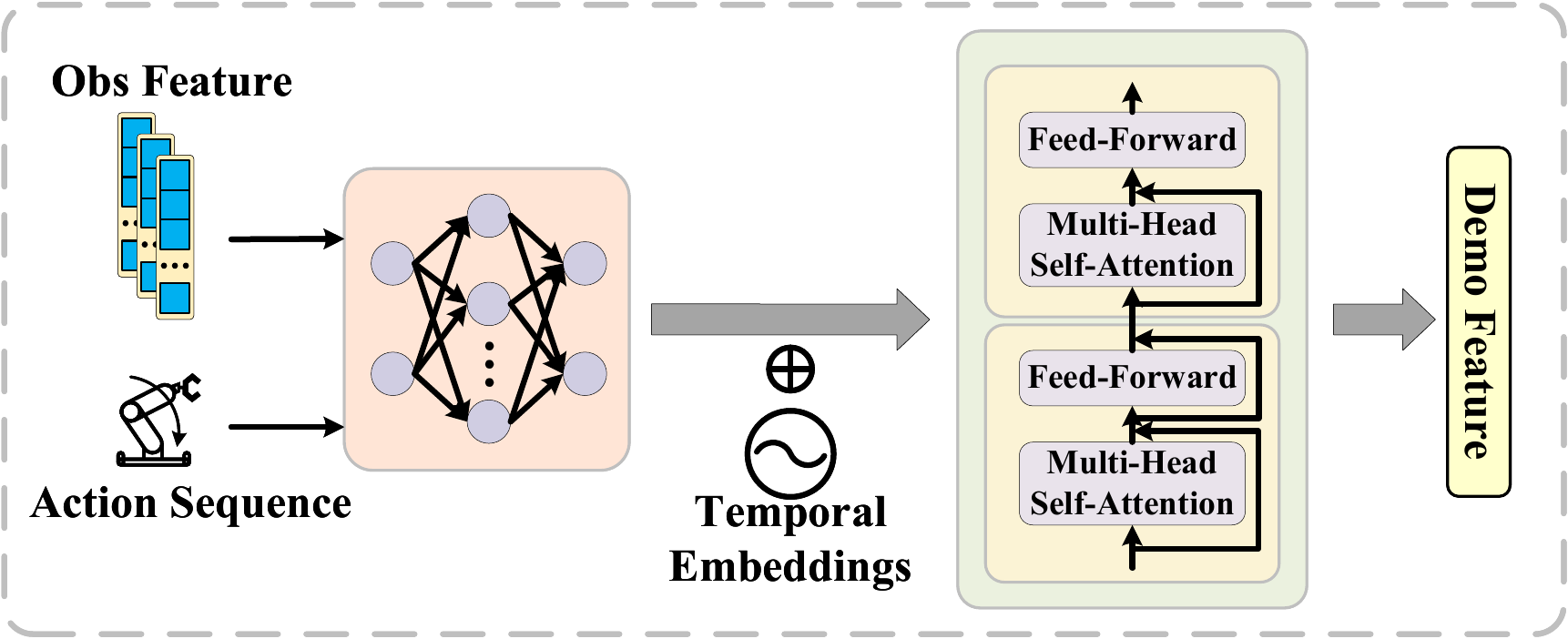}
	\caption{\textcolor{blue}{\textbf{VMT Encoding.} Observation features and expert actions are jointly embedded and encoded by a Transformer with learnable positional embeddings, preserving the full demonstration sequence.}}
	\label{fig:VMT}
	\vspace*{-0.4cm}
\end{figure}

\section{Method}

\textbf{Overview}. This section details the core design of DAMI. As shown in \Cref{fig:model}, \textcolor{blue}{the visual encoder processes current point clouds and robot states, frozen CLIP encodes language, and VMT encodes a complete reference demonstration. U2T fuses demonstration and text tokens with the observation--diffusion-timestep condition. TCFM uses this fusion in the first U-Net downsampling block, while other blocks retain the observation--timestep condition. Meta-training learns a shared initialization for adaptation with target-task demonstrations.}

\subsection{Problem Formulation}

We consider manipulation tasks $\mathcal{T}_i \sim p(\mathcal{T})$, each associated with a task-specific expert dataset $\mathcal{D}_{\mathcal{T}_i}$ containing ten complete expert episodes. \textcolor{blue}{Under the original sampling protocol}, a batch of observation--action windows is sampled from $\mathcal{D}_{\mathcal{T}_i}$ and split by index into a support batch $B_i^{sup}$ for inner-loop adaptation and a query batch $B_i^{qry}$ for the outer-loop update. A complete reference demonstration $\tau_i$ is obtained by randomly selecting one complete episode from the same task dataset. Thus, the demonstration, support batch, and query batch all originate from $\mathcal{D}_{\mathcal{T}_i}$\textcolor{blue}{. This original protocol does not enforce} episode-level isolation, \textcolor{blue}{so} the selected demonstration episode may also contribute windows to $B_i^{sup}$ or $B_i^{qry}$\textcolor{blue}{. Section IV-A details the sampling protocols and their scope}. At each control step, the policy $\pi_\theta$ takes the current observation $\mathbf{O}_t$ (a 3D point cloud and robot state), a text instruction $\ell_i$, and $\tau_i$ as inputs, and predicts an $H$-step action trajectory $\mathbf{A}_0 = [\mathbf{a}_t,\dots,\mathbf{a}_{t+H-1}]$.

\subsection{Diffusion-based Meta-Imitation Formulation}

DAMI adopts the Diffusion Policy paradigm \cite{chi2025diffusion} to model action trajectories. Given the clean action trajectory $\mathbf{A}_0$, the forward diffusion process gradually perturbs it into a noisy trajectory $\mathbf{A}_k$. The reverse process is implemented by a conditional U-Net that predicts the clean trajectory from the noisy sample, diffusion step, current observation feature $\mathbf{f}_{obs}$, demonstration tokens $\mathbf{H}_{demo}$, and text feature $\mathbf{f}_{text}$. Following DP3 \cite{Ze2024DP3}, we use the action-prediction parameterization:
\begin{equation}
	\mathcal{L}_{diff} =
	\mathbb{E}_{k,\mathbf{A}_0,\epsilon}
	\left[
	\left\|
	\mathbf{A}_0 -
		\hat{\mathbf{A}}_{0,\theta}(\mathbf{A}_k,k,\mathbf{f}_{obs},\mathbf{H}_{demo},\mathbf{f}_{text})
	\right\|^2
	\right].
	\label{eq:diff_loss}
\end{equation}

\textcolor{blue}{In \Cref{alg:train,alg:eval_ttft}, $\mathcal{L}_{diff}(w,B,d,\mathbf{f}_{text})$ denotes the diffusion loss in Equation~\eqref{eq:diff_loss}. All trainable encoders and U2T--TCFM conditions are recomputed using parameters $w$.}

\textcolor{blue}{We optimize the} diffusion objective \textcolor{blue}{using} the \textcolor{blue}{standard} full \textcolor{blue}{second-order} Model-Agnostic Meta-Learning (MAML) procedure\textcolor{blue}{~}\cite{finn2017model} \textcolor{blue}{to learn a shared policy initialization}. For a task $\mathcal{T}_i$, the inner loop adapts the current initialization $\theta$ on the support set:
\begin{equation}
	\theta'_i = \theta - \alpha \nabla_\theta \mathcal{L}_{diff}(\theta, B_i^{sup}),
	\label{eq:inner_loop}
\end{equation}
where $\alpha$ is the inner learning rate. At each outer step, the task batch $\mathcal{B}$ is constructed by enumerating all tasks assigned to the current GPU worker. The shared initialization is updated from the average query-set meta-gradient using AdamW:
\begin{equation}
	\begin{aligned}
		\bar{\mathbf{g}}_t &= \nabla_\theta \frac{1}{|\mathcal{B}|}
		\sum_{\mathcal{T}_i \in \mathcal{B}}
		\mathcal{L}_{diff}(\theta'_i, B_i^{qry}), \\
		\theta &\leftarrow \operatorname{AdamW}(\theta,\bar{\mathbf{g}}_t,\eta_t),
	\end{aligned}
	\label{eq:outer_loop}
\end{equation}
where $\eta_t$ follows a linear-warmup and cosine-decay schedule. We retain the computation graph through the inner-loop update, so the meta-gradient includes the second-order derivative terms of full MAML. During evaluation, base tasks are handled by direct inference using the EMA policy obtained during meta-training, whereas novel tasks are initialized from the same EMA policy, fine-tuned on few-shot demonstrations, and then evaluated with the adapted policy.

\subsection{The VMT Module}

\textcolor{blue}{We introduce the Visual-Motor Trajectory (VMT) module to encode the temporal sequence of visual features and expert actions in a complete reference demonstration, as illustrated in \Cref{fig:VMT}.} \textcolor{blue}{The design makes both the observed} visual \textcolor{blue}{information and the demonstrated control sequence available} for \textcolor{blue}{policy conditioning. These complementary inputs are intended to help distinguish} manipulation intents such as pushing, pulling, and pressing.
Formally, we consider a reference demonstration trajectory $\tau = \{(\mathbf{f}_{v,t}, \mathbf{a}_t)\}_{t=1}^T$, comprising \textcolor{blue}{observation} representations $\mathbf{f}_{v,t} \in \mathbb{R}^{D_v}$ \textcolor{blue}{that combine point-cloud and robot-state features} and \textcolor{blue}{expert} actions $\mathbf{a}_t \in \mathbb{R}^{D_a}$, where $T$ denotes the temporal horizon.

\textbf{Unified Modality Embedding.} \textcolor{blue}{We concatenate the observation feature} and \textcolor{blue}{expert} action \textcolor{blue}{at each time step and map them into} a shared \textcolor{blue}{token space through an MLP} $\phi_{\text{proj}}$:
\begin{equation}
	\mathbf{e}_t = \phi_{\text{proj}}([\mathbf{f}_{v,t} \oplus \mathbf{a}_t]),
\end{equation}
where $\oplus$ denotes concatenation, and $\mathbf{e}_t \in \mathbb{R}^{D_{model}}$ represents the unified token at step $t$. This results in a raw token sequence $\mathbf{E} = [\mathbf{e}_1, \dots, \mathbf{e}_T]$\textcolor{blue}{. VMT preserves sequence length without temporal downsampling or pooling}.

\textbf{Spatio-Temporal Representation Learning.} To capture trajectory-level evolution, we add learnable positional embeddings $\mathbf{P}$, implemented with an embedding layer, to $\mathbf{E}$ to retain sequential order. The sequence is then processed by a Transformer encoder to model global temporal dependencies:
\begin{equation}
	\mathbf{H}_{demo} = \text{Transformer}(\mathbf{E} + \mathbf{P}).
\end{equation}

The resulting token sequence $\mathbf{H}_{demo}$ \textcolor{blue}{provides demonstration} context \textcolor{blue}{for U2T fusion. VMT is optimized through the action-prediction objective in Equation~\eqref{eq:diff_loss} under MAML}.

\begin{algorithm*}[!t]
	\caption{DAMI Meta-Training Algorithm}
	\label{alg:train}
	\begin{algorithmic}[1]
		\REQUIRE Policy $\pi_\theta$, inner learning rate $\alpha$, outer optimizer $\mathcal{O}_{meta}$, EMA schedule $\{\gamma_t\}$, per-task datasets $\{\mathcal{D}_i\}$, \textcolor{blue}{sampling protocol $q$ (original or disjoint)}
		\STATE Load the frozen CLIP encoder, initialize $\theta_{\mathrm{ema}} \leftarrow \theta$
		\REPEAT
		\STATE Construct $\mathcal{B}$ from all tasks assigned to the current GPU worker, set $g_{total} \leftarrow 0$
		\FOR{each task $\mathcal{T}_i \in \mathcal{B}$}
		\IF{\textcolor{blue}{$q=\mathrm{disjoint}$ (additional ML10 comparisons and ablations)}}
		\STATE \textcolor{blue}{Partition the ten episode IDs afresh into disjoint sets $E_i^{ref},E_i^{sup},E_i^{qry}$ of sizes 1, 4, and 5}
		\STATE \textcolor{blue}{Load the complete reference $\tau_i$ from $E_i^{ref}$ and sample $B_i^{sup},B_i^{qry}$ only from their respective episode pools}
		\ELSE
		\STATE Sample a window batch $B_i\sim\mathcal{D}_i$, split it by index into $B_i^{sup}$ and $B_i^{qry}$
		\STATE \textcolor{blue}{Sample a complete reference trajectory $\tau_i$ uniformly from $\mathcal{D}_i$, allowing overlap with support or query episodes}
		\ENDIF
		\STATE $\mathbf{f}_{text} \leftarrow \mathrm{CLIP}(\ell_i)$
		\STATE \textcolor{blue}{$\theta'_i \leftarrow \theta-\alpha\nabla_\theta\mathcal{L}_{diff}(\theta,B_i^{sup},\tau_i,\mathbf{f}_{text})$, retaining the inner-update computation graph}
		\STATE \textcolor{blue}{$g_{total} \leftarrow g_{total}+\nabla_\theta\mathcal{L}_{diff}(\theta'_i,B_i^{qry},\tau_i,\mathbf{f}_{text})$} \hfill $\triangleright$ \textcolor{blue}{Full second-order meta-gradient}
		\ENDFOR
		\STATE $\theta \leftarrow \mathcal{O}_{meta}(\theta,g_{total}/|\mathcal{B}|)$ \hfill $\triangleright$ AdamW with warmup and cosine decay
		\STATE $\theta_{\mathrm{ema}} \leftarrow \gamma_t\theta_{\mathrm{ema}}+(1-\gamma_t)\theta$
		\UNTIL{the maximum number of epochs is reached}
		\STATE \textbf{return} $\pi_{\theta_{\mathrm{ema}}}$
	\end{algorithmic}
\end{algorithm*}

\begin{algorithm*}[!t]
	\caption{Novel-Task Few-Shot Adaptation and Evaluation Algorithm}
	\label{alg:eval_ttft}
	\begin{algorithmic}[1]
		\REQUIRE Pre-trained EMA policy $\pi_{\theta_{\mathrm{ema}}}$, target-task expert dataset $\mathcal{D}$, text instruction $\ell$, steps $N$, learning rate $\alpha$, weight decay $\lambda$, clipping threshold $c$, EMA \textcolor{blue}{decays} $\{\gamma_i\}$ \textcolor{blue}{($\gamma_i=0.75$ for ML10)}
		\STATE $\pi_\phi \leftarrow \mathrm{Copy}(\pi_{\theta_{\mathrm{ema}}})$, $\phi_{\mathrm{ema}} \leftarrow \phi$
		\STATE \textcolor{blue}{For ML10, inherit checkpoint normalizers without refitting and include their affine parameters in adaptation updates}
		\STATE $\mathbf{f}_{text} \leftarrow \mathrm{CLIP}(\ell)$
		\FOR{$i=1$ to $N$}
		\STATE \textcolor{blue}{Sample reference trajectory $d$ and observation--action windows $B$ from the same $\mathcal{D}$, allowing episode overlap}
		\STATE \textcolor{blue}{$\mathcal{L} \leftarrow \mathcal{L}_{diff}(\phi,B,d,\mathbf{f}_{text})$}
		\STATE $\phi \leftarrow \mathrm{AdamW}(\phi,\mathrm{Clip}(\nabla_\phi\mathcal{L},c),\alpha,\lambda)$
		\STATE $\phi_{\mathrm{ema}} \leftarrow \gamma_i\phi_{\mathrm{ema}}+(1-\gamma_i)\phi$
		\ENDFOR
\STATE \textcolor{blue}{For each Meta-World rollout, sample $d_{eval}\sim\mathcal{D}$ and hold it fixed within that rollout}
\STATE \textcolor{blue}{For each RLBench evaluation with a given task, checkpoint, and evaluation seed, sample one $d_{eval}\sim\mathcal{D}$ and reuse it across its 20 episodes}
		\STATE \textcolor{blue}{Execute $\pi_{\phi_{\mathrm{ema}}}$ with these references and compute success rate as successful rollouts divided by total rollouts}
		\STATE \textbf{return} \textcolor{blue}{success rate}
	\end{algorithmic}
\end{algorithm*}

\subsection{The U2T Module}

\textbf{Cross-Modal Semantic Alignment.} To harmonize an unstructured text instruction, a structured reference demonstration, and the current observation, we propose the Unpaired Unified Task encoder. Here, ``unpaired'' \textcolor{blue}{means} that the reference demonstration and current observation \textcolor{blue}{belong to the same task but} need not be \textcolor{blue}{frame-wise} aligned.

The frozen CLIP encoder \cite{radford2021learning} produces the text feature $\mathbf{f}_{text}$, and VMT produces the demonstration tokens $\mathbf{H}_{demo}$. We use modality-specific affine projections to align these two fusion inputs:
\begin{equation}
\mathbf{e}_{m} = \mathbf{W}_m \cdot \text{LN}(\mathbf{f}_m) + \mathbf{b}_m,
\quad m \in \{demo,text\},
\end{equation}
where $\mathbf{f}_{demo}=\mathbf{H}_{demo}$ and $\text{LN}(\cdot)$ denotes LayerNorm. The current observation feature $\mathbf{f}_{obs}$ is not passed through this modality-projection branch. \textcolor{blue}{Within each update, samples from the same task share a reference demonstration and instruction. The full demonstration sequence is retained without pooling.}

\textbf{Observation--Timestep-Conditioned Fusion.} At diffusion step $k$, the timestep embedding $\mathbf{e}_k=\mathrm{TimeEmb}(k)$ is concatenated with the current observation feature\textcolor{blue}{:}
\begin{equation}
\textcolor{blue}{\mathbf{h}_k=[\mathbf{e}_k\oplus\mathbf{f}_{obs}].}
\label{eq:obs_time_condition}
\end{equation}
\textcolor{blue}{For each residual block $r$ of the first downsampling stage, a block-specific nonlinear projection forms an observation--timestep token $\mathbf{o}_{k,r}$.} Distinct learnable positional embeddings \textcolor{blue}{are added to} the demonstration\textcolor{blue}{, text,} and \textcolor{blue}{observation--timestep} tokens. The U2T input and resulting condition are
\begin{equation}
\begin{aligned}
\mathbf{Z}_{\textcolor{blue}{r}}&=[\mathbf{e}_{demo}+\mathbf{P}_{demo}\,||\,
\mathbf{e}_{text}+\mathbf{P}_{text}\\
&\qquad ||\,\textcolor{blue}{\mathbf{o}_{k,r}+\mathbf{P}_{obs}}],\\
\textcolor{blue}{\mathbf{c}_r}&=\mathrm{U2T}(\mathbf{Z}_{\textcolor{blue}{r}}).
\end{aligned}
\label{eq:u2t_condition}
\end{equation}
\textcolor{blue}{Here,} $||$ denotes sequence concatenation. \textcolor{blue}{U2T prepends learnable query tokens to $\mathbf{Z}_r$, fuses the sequence with a Transformer, and reads the first query output as $\mathbf{c}_r$.}

\subsection{Task-Conditioned Feature Modulation}

We implement the denoising network as a modified 1D conditional U-Net with residual convolutional blocks. \textcolor{blue}{We} apply U2T fusion only at the first downsampling block. The remaining blocks use the observation--timestep condition $\mathbf{h}_k$ from \Cref{eq:obs_time_condition}.

\textcolor{blue}{For each residual block $r$ in this stage, an independent linear layer maps the fused condition to FiLM parameters:}
{
\color{blue}
\begin{equation}
[\boldsymbol{\gamma}_r\oplus\boldsymbol{\beta}_r]
=\mathbf{W}_{f,r}\mathbf{c}_r+\mathbf{b}_{f,r}.
\label{eq:tcfm_condition}
\end{equation}
}%
\textcolor{blue}{Splitting the output equally gives channel-wise scales $\boldsymbol{\gamma}_r$ and offsets $\boldsymbol{\beta}_r$, which are broadcast over the action-time dimension of} the feature \textcolor{blue}{$\mathbf{X}_r$:}
{
\color{blue}
\begin{equation}
Y_{r,b,c,\tau}=\gamma_{r,b,c}X_{r,b,c,\tau}+\beta_{r,b,c}.
\label{eq:tcfm}
\end{equation}
}%
\textcolor{blue}{Each sample and channel has its own scale and bias, shared across temporal positions. The scale is applied directly, without adding one. The residual blocks} at \textcolor{blue}{this stage share U2T parameters but have separate observation--timestep projections and FiLM-generating layers}. \textcolor{blue}{Each calls U2T and generates its own coefficients.}

\subsection{Training and Adaptation Procedures}

\textbf{Meta-training.} The complete procedure is summarized in \Cref{alg:train}. At each outer step, every GPU worker constructs its task batch by enumerating all tasks assigned to that worker rather than randomly drawing tasks from $p(\mathcal{T})$. Each task has one expert dataset $\mathcal{D}_i$ containing ten complete expert episodes. \textcolor{blue}{Support and query batches are formed from} observation--action windows \textcolor{blue}{sampled} from this dataset. The reference demonstration is sampled from the same task dataset by randomly selecting one complete episode and loading the \textcolor{blue}{complete trajectory}. VMT encodes the demonstration, the frozen CLIP encoder processes the text instruction, and the visual encoder extracts the current observation feature. U2T fuses \textcolor{blue}{demonstration, language, and observation--timestep information to condition} the \textcolor{blue}{diffusion policy}. The inner loop performs one action-prediction gradient update on all trainable policy parameters. The complete query forward pass, including VMT, the visual encoder, U2T, and the U-Net, is recomputed using $\theta'_i$\textcolor{blue}{. Only} the frozen CLIP feature is reused. Full-MAML gradients are averaged across the task batch and applied with AdamW under a linear-warmup and cosine-decay learning-rate schedule. An EMA copy of the policy is maintained with a step-dependent decay schedule.

\textbf{Novel-task Adaptation and Evaluation.} The few-shot adaptation procedure is summarized in \Cref{alg:eval_ttft}. For \textcolor{blue}{each novel task, adaptation uses its} expert dataset $\mathcal{D}$. \textcolor{blue}{The same $\mathcal{D}$ supplies the fine-tuning windows and complete reference demonstrations. In ML10, adaptation inherits checkpoint normalization statistics without refitting, while affine parameters remain in the adaptation optimizer.} Each adaptation iteration samples an observation--action batch and a complete reference trajectory from $\mathcal{D}$, rebuilds the \textcolor{blue}{policy} conditions, and updates the policy using AdamW with weight decay and gradient clipping. \textcolor{blue}{An} EMA copy of the adapted policy is maintained \textcolor{blue}{and used for evaluation, with a fixed decay of $0.75$ in ML10}. \textcolor{blue}{Base tasks use direct inference, with reference demonstrations from their training-task datasets and no adaptation support set.}

\textcolor{blue}{\textbf{Reference Sampling during Evaluation.}} For Meta-World, one complete \textcolor{blue}{reference} demonstration is randomly selected at the start of each rollout \textcolor{blue}{and} held fixed throughout that rollout. RLBench samples one \textcolor{blue}{reference for each evaluation run with a given task, checkpoint, and evaluation seed} and reuses it across \textcolor{blue}{twenty} episodes. \textcolor{blue}{Both benchmarks resample reference demonstrations at each novel-task adaptation update.}

\section{Experiment}
\subsection{Experiment Setup}
\label{sec:experiment_setup}

\begin{table*}[t]
	\centering
	\caption{Key Hyperparameters Used in Our Experiments.}
	\label{tab:hyperparams}
	\footnotesize
	\setlength{\tabcolsep}{4pt}
	\renewcommand{\arraystretch}{1.05}
	\begin{tabular*}{\textwidth}{@{\extracolsep{\fill}}cccccc}
		\toprule
		\multicolumn{2}{c}{\textbf{Training and Data}} &
		\multicolumn{2}{c}{\textbf{Optimization}} &
		\multicolumn{2}{c}{\textbf{Architecture and Adaptation}} \\
		\cmidrule(lr){1-2}\cmidrule(lr){3-4}\cmidrule(lr){5-6}
		\textbf{Hyperparameter} & \textbf{Value} &
		\textbf{Hyperparameter} & \textbf{Value} &
		\textbf{Hyperparameter} & \textbf{Value} \\
		\midrule
		Training Epochs & 3,000 & Outer Optimizer & AdamW & Point Feature Dim & 64 \\
\textcolor{blue}{\makecell[c]{Outer-loop Updates\\per Epoch}} & 18 & Outer Learning Rate & $1.0 \times 10^{-4}$ & \textcolor{blue}{VMT Token Dim} & 128 \\
		Prediction Horizon & 16 & Optimizer Betas & [0.95, 0.999] & VMT Attention Heads & 4 \\
		Action Pred. Step & 8 & Weight Decay & $1.0 \times 10^{-6}$ & VMT Layers & 2 \\
		Observation Step & 2 & Inner Optimizer & SGD & U2T Attention Heads & 8 \\
		Diff. Training Steps & 100 & Inner Learning Rate & $1.0 \times 10^{-4}$ & \textcolor{blue}{\makecell[c]{U2T demonstration\\positions (Meta-World)}} & 200 \\
		Diff. Inference Steps & 10 & Inner Loop Steps & 1 & U2T Pos. Emb. (Real) & 500 \\
		Support / Episodes per Task & 20 / 10 & LR Scheduler & Warmup + Cosine & Diff. Step Embed Dim & 128 \\
		Query Set Size & 88 & \makecell[c]{\textcolor{blue}{Meta-training}\\EMA Power / Max Decay} & 0.75 / 0.9999 & FT Vision LR & $1.0 \times 10^{-5}$ \\
		Total Batch Size & 108 & FT Diffusion LR & $3.0 \times 10^{-5}$ & FT Other LR & $1.0 \times 10^{-4}$ \\
		\bottomrule
	\end{tabular*}
	\vspace{-0.1cm}
\end{table*}

\begin{table*}[!t]
\centering
\caption{Success Rates (\%) on Base Meta-World Tasks.}
\label{seen-result-table}
\scriptsize
\setlength{\tabcolsep}{1pt}\renewcommand{\arraystretch}{1.05}
\begin{tabular*}{\textwidth}{@{\extracolsep{\fill}}c ccccccccc}
\toprule
& \multicolumn{3}{c}{ML10 Tasks} & \multicolumn{4}{c}{ML45 Tasks} & \multicolumn{2}{c}{Average} \\ \cmidrule(lr){2-4}\cmidrule(lr){5-8}\cmidrule(lr){9-10}
\textcolor{blue}{Method} & E(5) & M(3) & H(2) & E(26) & M(9) & H(5) & VH(5) & ML10 & ML45 \\
\midrule
\textcolor{blue}{MAML-TRPO} & \mbox{74.40\textcolor{blue}{\,$\pm$\,7.67}} & \mbox{1.33\textcolor{blue}{\,$\pm$\,1.15}} & \mbox{10.00\textcolor{blue}{\,$\pm$\,1.32}} & \mbox{13.23\textcolor{blue}{\,$\pm$\,2.97}} & \mbox{1.41\textcolor{blue}{\,$\pm$\,1.51}} & \mbox{0.00\textcolor{blue}{\,$\pm$\,0.00}} & \mbox{0.00\textcolor{blue}{\,$\pm$\,0.00}} & \mbox{39.60\textcolor{blue}{\,$\pm$\,4.36}} & \mbox{7.93\textcolor{blue}{\,$\pm$\,1.82}} \\
DP3 & \textcolor{blue}{\mbox{\underline{87.13}\,$\pm$\,0.83}} & \mbox{80.11\textcolor{blue}{\,$\pm$\,2.91}} & \mbox{\textbf{72.83}\textcolor{blue}{\,$\pm$\,3.82}} & \mbox{59.47\textcolor{blue}{\,$\pm$\,0.79}} & \mbox{34.93\textcolor{blue}{\,$\pm$\,2.09}} & \mbox{23.60\textcolor{blue}{\,$\pm$\,2.95}} & \mbox{\underline{29.40}\textcolor{blue}{\,$\pm$\,0.53}} & \textcolor{blue}{\mbox{\underline{82.17}\,$\pm$\,0.75}} & \mbox{47.24\textcolor{blue}{\,$\pm$\,1.01}} \\
Mamba & \mbox{85.07\textcolor{blue}{\,$\pm$\,1.70}} & \mbox{76.22\textcolor{blue}{\,$\pm$\,4.19}} & \mbox{33.83\textcolor{blue}{\,$\pm$\,2.52}} & \mbox{\underline{64.18}\textcolor{blue}{\,$\pm$\,0.89}} & \mbox{\underline{36.48}\textcolor{blue}{\,$\pm$\,2.06}} & \mbox{\underline{32.80}\textcolor{blue}{\,$\pm$\,4.89}} & \mbox{25.87\textcolor{blue}{\,$\pm$\,1.80}} & \mbox{72.17\textcolor{blue}{\,$\pm$\,1.94}} & \mbox{\underline{50.90}\textcolor{blue}{\,$\pm$\,0.84}} \\
FreqPolicy & \mbox{84.33\textcolor{blue}{\,$\pm$\,3.06}} & \mbox{\underline{82.11}\textcolor{blue}{\,$\pm$\,3.29}} & \mbox{\underline{69.83}\textcolor{blue}{\,$\pm$\,12.91}} & \mbox{57.88\textcolor{blue}{\,$\pm$\,0.81}} & \mbox{32.63\textcolor{blue}{\,$\pm$\,3.99}} & \mbox{26.00\textcolor{blue}{\,$\pm$\,3.61}} & \mbox{27.07\textcolor{blue}{\,$\pm$\,6.40}} & \mbox{80.77\textcolor{blue}{\,$\pm$\,0.90}} & \mbox{45.87\textcolor{blue}{\,$\pm$\,1.40}} \\
FlowPolicy & \mbox{83.73\textcolor{blue}{\,$\pm$\,0.70}} & \mbox{74.44\textcolor{blue}{\,$\pm$\,4.67}} & \mbox{69.17\textcolor{blue}{\,$\pm$\,5.80}} & \mbox{61.21\textcolor{blue}{\,$\pm$\,0.76}} & \mbox{28.81\textcolor{blue}{\,$\pm$\,0.55}} & \mbox{15.80\textcolor{blue}{\,$\pm$\,3.47}} & \mbox{26.73\textcolor{blue}{\,$\pm$\,2.32}} & \mbox{78.03\textcolor{blue}{\,$\pm$\,0.32}} & \mbox{45.85\textcolor{blue}{\,$\pm$\,1.01}} \\
\rowcolor{gray!15}\textbf{Ours (DAMI)} & \mbox{\textbf{90.27}\textcolor{blue}{\,$\pm$\,1.42}} & \mbox{\textbf{91.78}\textcolor{blue}{\,$\pm$\,4.00}} & \mbox{\textbf{72.83}\textcolor{blue}{\,$\pm$\,5.39}} & \mbox{\textbf{87.19}\textcolor{blue}{\,$\pm$\,1.00}} & \mbox{\textbf{71.04}\textcolor{blue}{\,$\pm$\,1.88}} & \mbox{\textbf{61.60}\textcolor{blue}{\,$\pm$\,2.84}} & \mbox{\textbf{68.60}\textcolor{blue}{\,$\pm$\,4.51}} & \mbox{\textbf{87.23}\textcolor{blue}{\,$\pm$\,2.02}} & \mbox{\textbf{79.05}\textcolor{blue}{\,$\pm$\,1.24}} \\
\bottomrule\end{tabular*}

\par\smallskip\begin{minipage}{\textwidth}\fontsize{7}{8.1}\selectfont\raggedright \textcolor{blue}{Mean $\pm$ SD over three evaluation seeds. Bold and underline indicate first and second ranks, including ties.}\end{minipage}
\par\vspace{-0.1cm}
\end{table*}

\begin{table*}[!t]
\centering
\caption{Success Rates (\%) on Novel Meta-World Tasks.}
\label{unseen-result-table}
\fontsize{6.5}{7.8}\selectfont
\setlength{\tabcolsep}{1pt}\renewcommand{\arraystretch}{1.05}
\begin{tabular*}{\textwidth}{@{\extracolsep{\fill}}c cccccccccccc}
\toprule
& \multicolumn{5}{c}{ML10 Tasks} & \multicolumn{5}{c}{ML45 Tasks} & \multicolumn{2}{c}{Average} \\ \cmidrule(lr){2-6}\cmidrule(lr){7-11}\cmidrule(lr){12-13}
\textcolor{blue}{Method} & \makecell{Door\\Close} & \makecell{Drawer\\Open} & \makecell{Lever\\Pull} & \makecell{Sweep\\Into} & \makecell{Shelf\\Place} & \makecell{Door\\Lock} & \makecell{Door\\Unlock} & \makecell{Bin\\Picking} & \makecell{Box\\Close} & \makecell{Hand\\Insert} & ML10 & ML45 \\
\midrule
\textcolor{blue}{MAML-TRPO} & \mbox{\underline{94.00}\textcolor{blue}{\,$\pm$\,10.39}} & \mbox{22.67\textcolor{blue}{\,$\pm$\,19.66}} & \mbox{5.33\textcolor{blue}{\,$\pm$\,5.03}} & \mbox{\underline{24.67}\textcolor{blue}{\,$\pm$\,5.77}} & \mbox{0.00\textcolor{blue}{\,$\pm$\,0.00}} & \mbox{\underline{42.67}\textcolor{blue}{\,$\pm$\,24.38}} & \mbox{7.67\textcolor{blue}{\,$\pm$\,9.29}} & \mbox{0.33\textcolor{blue}{\,$\pm$\,0.58}} & \mbox{1.67\textcolor{blue}{\,$\pm$\,1.53}} & \mbox{\underline{10.33}\textcolor{blue}{\,$\pm$\,15.31}} & \mbox{29.33\textcolor{blue}{\,$\pm$\,6.63}} & \mbox{12.53\textcolor{blue}{\,$\pm$\,7.94}} \\
DP3 & \mbox{\textbf{100.00}\textcolor{blue}{\,$\pm$\,0.00}} & \mbox{70.33\textcolor{blue}{\,$\pm$\,27.43}} & \mbox{\underline{44.67}\textcolor{blue}{\,$\pm$\,4.51}} & \mbox{16.67\textcolor{blue}{\,$\pm$\,6.03}} & \mbox{\textbf{12.67}\textcolor{blue}{\,$\pm$\,5.03}} & \mbox{\textbf{51.00}\textcolor{blue}{\,$\pm$\,25.24}} & \mbox{\underline{94.00}\textcolor{blue}{\,$\pm$\,1.73}} & \mbox{9.33\textcolor{blue}{\,$\pm$\,7.64}} & \mbox{\textbf{26.67}\textcolor{blue}{\,$\pm$\,7.23}} & \mbox{6.67\textcolor{blue}{\,$\pm$\,4.93}} & \mbox{\underline{48.87}\textcolor{blue}{\,$\pm$\,7.20}} & \mbox{\textbf{37.53}\textcolor{blue}{\,$\pm$\,4.24}} \\
Mamba & \mbox{\textbf{100.00}\textcolor{blue}{\,$\pm$\,0.00}} & \mbox{32.33\textcolor{blue}{\,$\pm$\,15.01}} & \mbox{27.67\textcolor{blue}{\,$\pm$\,6.11}} & \textcolor{blue}{\mbox{9.33\,$\pm$\,4.04}} & \mbox{10.00\textcolor{blue}{\,$\pm$\,3.00}} & \mbox{22.67\textcolor{blue}{\,$\pm$\,11.85}} & \mbox{34.67\textcolor{blue}{\,$\pm$\,16.20}} & \mbox{9.67\textcolor{blue}{\,$\pm$\,1.15}} & \mbox{12.67\textcolor{blue}{\,$\pm$\,6.81}} & \mbox{5.33\textcolor{blue}{\,$\pm$\,4.51}} & \textcolor{blue}{\mbox{35.87\,$\pm$\,4.56}} & \mbox{\underline{17.00}\textcolor{blue}{\,$\pm$\,4.42}} \\
FreqPolicy & \mbox{93.00\textcolor{blue}{\,$\pm$\,12.12}} & \mbox{\underline{98.00}\textcolor{blue}{\,$\pm$\,3.46}} & \mbox{18.67\textcolor{blue}{\,$\pm$\,1.15}} & \mbox{15.67\textcolor{blue}{\,$\pm$\,3.21}} & \mbox{7.00\textcolor{blue}{\,$\pm$\,4.58}} & \mbox{11.33\textcolor{blue}{\,$\pm$\,6.66}} & \mbox{6.00\textcolor{blue}{\,$\pm$\,5.00}} & \mbox{\underline{11.00}\textcolor{blue}{\,$\pm$\,6.56}} & \mbox{\underline{25.33}\textcolor{blue}{\,$\pm$\,8.62}} & \mbox{\textbf{13.67}\textcolor{blue}{\,$\pm$\,3.06}} & \mbox{46.47\textcolor{blue}{\,$\pm$\,1.33}} & \mbox{13.47\textcolor{blue}{\,$\pm$\,0.76}} \\
FlowPolicy & \mbox{\textbf{100.00}\textcolor{blue}{\,$\pm$\,0.00}} & \mbox{13.33\textcolor{blue}{\,$\pm$\,3.06}} & \mbox{17.67\textcolor{blue}{\,$\pm$\,11.15}} & \mbox{6.67\textcolor{blue}{\,$\pm$\,3.06}} & \mbox{0.00\textcolor{blue}{\,$\pm$\,0.00}} & \mbox{15.00\textcolor{blue}{\,$\pm$\,15.52}} & \mbox{22.00\textcolor{blue}{\,$\pm$\,13.45}} & \mbox{9.00\textcolor{blue}{\,$\pm$\,9.54}} & \mbox{13.33\textcolor{blue}{\,$\pm$\,13.05}} & \mbox{6.00\textcolor{blue}{\,$\pm$\,4.36}} & \mbox{27.53\textcolor{blue}{\,$\pm$\,1.94}} & \mbox{13.07\textcolor{blue}{\,$\pm$\,4.39}} \\
\rowcolor{gray!15}\textbf{Ours (DAMI)} & \mbox{\textbf{100.00}\textcolor{blue}{\,$\pm$\,0.00}} & \mbox{\textbf{100.00}\textcolor{blue}{\,$\pm$\,0.00}} & \mbox{\textbf{45.33}\textcolor{blue}{\,$\pm$\,18.45}} & \mbox{\textbf{28.00}\textcolor{blue}{\,$\pm$\,14.73}} & \mbox{\underline{10.67}\textcolor{blue}{\,$\pm$\,1.15}} & \mbox{39.33\textcolor{blue}{\,$\pm$\,27.15}} & \mbox{\textbf{98.67}\textcolor{blue}{\,$\pm$\,1.15}} & \mbox{\textbf{13.67}\textcolor{blue}{\,$\pm$\,10.07}} & \mbox{22.33\textcolor{blue}{\,$\pm$\,21.03}} & \mbox{\textbf{13.67}\textcolor{blue}{\,$\pm$\,1.15}} & \mbox{\textbf{56.80}\textcolor{blue}{\,$\pm$\,2.40}} & \mbox{\textbf{37.53}\textcolor{blue}{\,$\pm$\,4.22}} \\
\bottomrule\end{tabular*}

\par\smallskip\begin{minipage}{\textwidth}\fontsize{7}{8.1}\selectfont\raggedright \textcolor{blue}{Mean $\pm$ SD over three evaluation seeds. Bold and underline indicate first and second ranks, including ties.}\end{minipage}
\par\vspace{-0.1cm}
\end{table*}

\begin{table*}[!t]
\centering
\caption{\textcolor{blue}{Per-Task} Success Rates (\%) on Novel RLBench FS25 Tasks.}
\label{tab:rlbench_fs25}
\begingroup\footnotesize
\setlength{\tabcolsep}{4pt}\renewcommand{\arraystretch}{1.1}
\begin{tabular*}{\textwidth}{@{\extracolsep{\fill}}cccccc}
\toprule
\textcolor{blue}{Task} & DP3 & Mamba & FreqPolicy & FlowPolicy & \textcolor{blue}{DAMI} \\
\midrule
\textcolor{blue}{Light Bulb Out} & \textcolor{blue}{$0.00 \pm 0.00$} & \textcolor{blue}{$0.00 \pm 0.00$} & \textcolor{blue}{$0.00 \pm 0.00$} & \textcolor{blue}{$0.00 \pm 0.00$} & \textcolor{blue}{$0.00 \pm 0.00$} \\
\textcolor{blue}{TV On} & \textcolor{blue}{$0.00 \pm 0.00$} & \textcolor{blue}{$0.00 \pm 0.00$} & \textcolor{blue}{$0.00 \pm 0.00$} & \textcolor{blue}{$0.00 \pm 0.00$} & \textcolor{blue}{$0.00 \pm 0.00$} \\
\textcolor{blue}{Open Oven} & \textcolor{blue}{$0.00 \pm 0.00$} & \textcolor{blue}{$0.00 \pm 0.00$} & \textcolor{blue}{$0.00 \pm 0.00$} & \textcolor{blue}{$0.00 \pm 0.00$} & \textcolor{blue}{$\mathbf{1.00 \pm 1.00}$} \\
\textcolor{blue}{Open Door} & \textcolor{blue}{$49.67 \pm 3.79$} & \textcolor{blue}{$0.00 \pm 0.00$} & \textcolor{blue}{$6.00 \pm 1.73$} & \textcolor{blue}{$1.00 \pm 0.00$} & \textcolor{blue}{$\mathbf{53.00 \pm 4.00}$} \\
\textcolor{blue}{Take Item Out of Drawer} & \textcolor{blue}{$0.00 \pm 0.00$} & \textcolor{blue}{$0.00 \pm 0.00$} & \textcolor{blue}{$0.00 \pm 0.00$} & \textcolor{blue}{$0.00 \pm 0.00$} & \textcolor{blue}{$0.00 \pm 0.00$} \\
\midrule
\textcolor{blue}{\textbf{Five-task average}} & $9.93 \mathbin{\textcolor{blue}{\pm}} \textcolor{blue}{0.76}$ & $0.00 \mathbin{\textcolor{blue}{\pm}} \textcolor{blue}{0.00}$ & $1.20 \mathbin{\textcolor{blue}{\pm}} \textcolor{blue}{0.35}$ & $0.20 \mathbin{\textcolor{blue}{\pm}} \textcolor{blue}{0.00}$ & $\mathbf{10.80 \mathbin{\textcolor{blue}{\pm}} \textcolor{blue}{0.92}}$ \\
\bottomrule
\end{tabular*}
\par\smallskip\begin{minipage}{\textwidth}\scriptsize\raggedright
\textcolor{blue}{Values are means $\pm$ sample SDs across evaluation seeds. Each task--seed score averages the five highest checkpoint success rates after 200 adaptation steps. The overall score averages tasks within each seed before computing the cross-seed mean and sample SD. Bold indicates the highest nonzero mean in each row. Zero denotes no successful rollouts in the recorded evaluations.}
\end{minipage}\endgroup
\end{table*}



\textbf{Simulations.} \textcolor{blue}{We evaluate base-task direct inference and novel-task few-shot adaptation on Meta-World} ML10 and ML45\textcolor{blue}{~}\cite{yu2020meta} and \textcolor{blue}{RLBench} FS25\textcolor{blue}{~}\cite{james2020rlbench}. \textcolor{blue}{The three settings contain 10/5, 45/5, and 25/5 base/held-out novel tasks, respectively.}

\textbf{Expert \textcolor{blue}{Demonstrations}.} \textcolor{blue}{For Meta-World, we follow DP3~\cite{Ze2024DP3} and use task-specific scripted controllers to collect expert trajectories. RLBench demonstrations are generated using task waypoints and motion planning. All imitation-learning methods use the same datasets, with ten complete demonstrations per task.}

\begin{figure}[t]
	\centering
	\includegraphics[width=\columnwidth]{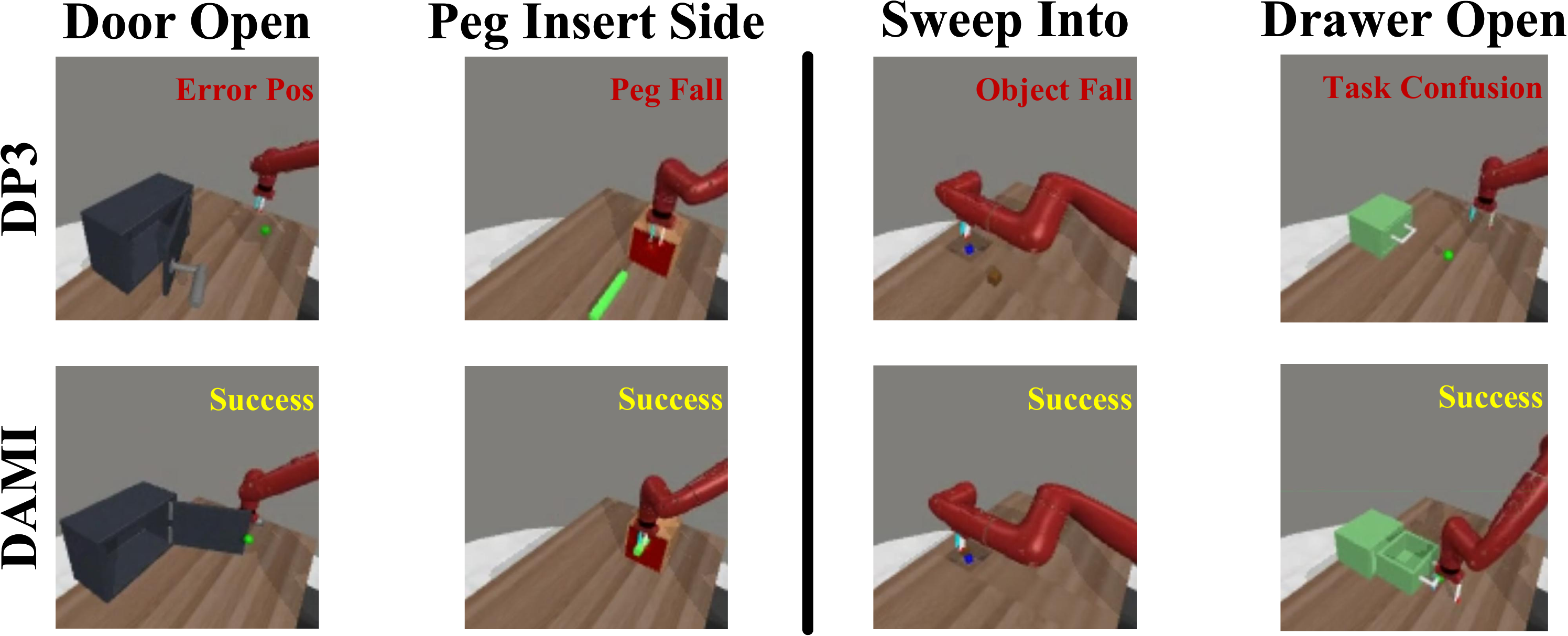}
	\caption{\textbf{\textcolor{blue}{Selected Executions} on Meta-World.} DP3 (top) and DAMI (bottom) \textcolor{blue}{use base-task direct inference} (\textcolor{blue}{Door Open, Peg Insert Side}) and \textcolor{blue}{200-step} novel-task adaptation (\textcolor{blue}{Sweep Into, Drawer Open}). \textcolor{blue}{Success/failure is judged by simulator criteria over complete rollouts. Examples illustrate behavior, not frequency.}}
	\label{fig:qualitative}
	\vspace{-0.5cm}
\end{figure}

\textbf{Baselines.} We \textcolor{blue}{compare} DAMI \textcolor{blue}{with} DP3\textcolor{blue}{~\cite{Ze2024DP3}}, FlowPolicy~\cite{zhang2025flowpolicy}, FreqPolicy~\cite{zhong2026freqpolicy}, and Mamba Policy~\cite{cao2025mamba}, \textcolor{blue}{jointly trained on multitask demonstrations. All four and their MAML variants condition on point clouds and robot states, without language, reference demonstrations, or task IDs. We train and evaluate Meta-World's MAML-TRPO~\cite{yu2020meta} on our ML10/ML45 splits and MAML~\cite{finn2017model} variants of all four baselines on ML10 (\Cref{tab:ml10_maml_summary}), denoted by ``MAML-''.}

\textcolor{blue}{On ML10, input-matched DP3 and MAML-DP3 controls share DAMI's current observations, language, and complete 200-frame references containing point clouds, robot states, and expert actions (\Cref{tab:ml10_matched_conditioning}). Using BC and MAML, respectively, they mean-pool per-frame MLP reference encodings and concatenate them with language and observation features for standard DP3 FiLM~\cite{perez2018film}, without VMT, U2T, or TCFM. Additional BC controls pair current observations with either language or a reference. A VMT variant replaces U2T and TCFM with pooling, concatenation, and an MLP feeding DP3 FiLM.}

\textcolor{blue}{\textbf{Training-Protocol Scope.} Original benchmark and MAML comparisons retain their protocols and DAMI results. Additional ML10 comparisons in \Cref{tab:episode_disjoint,tab:ml10_matched_conditioning,tab:ml10_component_configurations,tab:ml10_vmt_targeted} use episode-disjoint training and matched full-DAMI references. Each update repartitions each task's ten trajectories into one reference, four support, and five query trajectories, sampling windows within each pool. Configurations share the training seed and use 54,000 AdamW updates with EMA. MAML uses full second-order gradients and one SGD inner update. BC weights support/query diffusion losses by window counts. Novel-task adaptation samples references and support windows from a shared pool. Budget and reference-selection experiments reuse original models, varying only the corresponding test-time factors.}

\textbf{Evaluation Protocols.} \textcolor{blue}{Simulation results use evaluation seeds 0, 1, and 2, except the seven structural component configurations, which share one evaluation seed. Following DP3~\cite{Ze2024DP3}, imitation-learning methods are evaluated every 200 epochs with 20 rollouts per task. Success rate is the fraction satisfying the task criterion. Base tasks use direct inference, and novel tasks use 200 fine-tuning steps. DAMI uses EMA checkpoints for inference and adaptation initialization (\Cref{alg:eval_ttft}).}

\textcolor{blue}{Top-5 reporting averages the five highest checkpoint success rates separately for each task and evaluation seed. Category and overall scores average tasks within each seed. We then report cross-seed means and sample standard deviations (SDs).}

\begin{figure*}
	\centering
	\includegraphics[width=\textwidth]{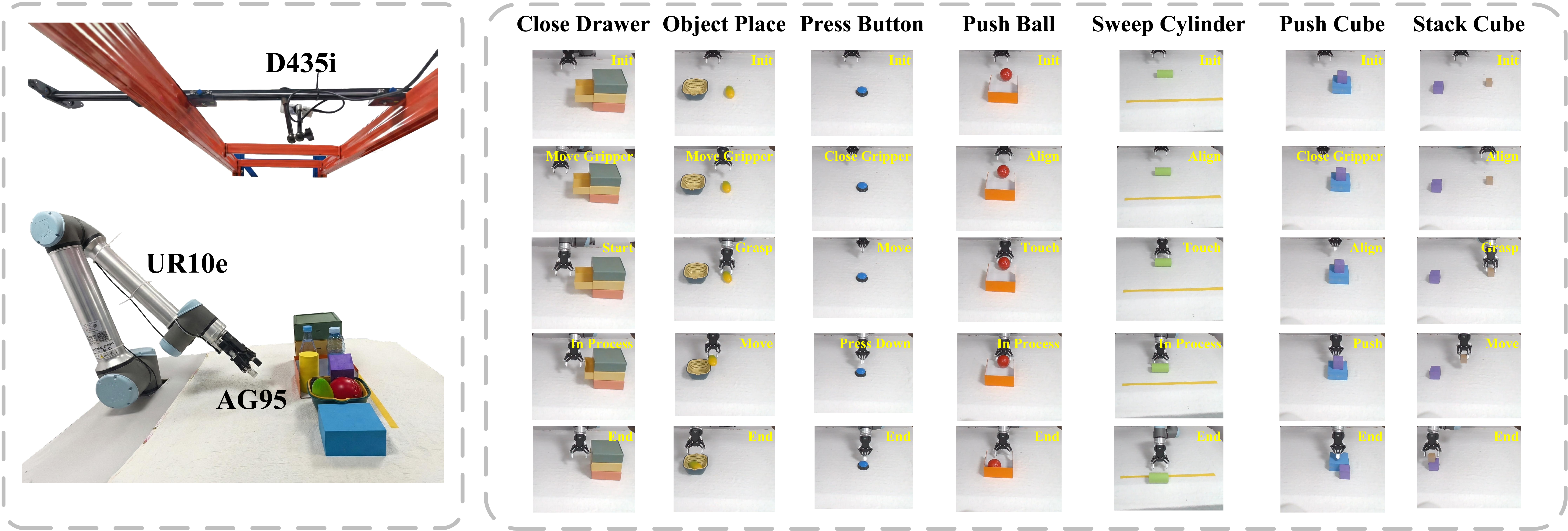}
	
	\caption{\textbf{\textcolor{blue}{Real-World Platform} and \textcolor{blue}{Task Examples}.} The left panel shows the UR10e manipulator\textcolor{blue}{,} AG95 gripper, \textcolor{blue}{and RealSense D435i camera. The right panels illustrate execution stages for five base tasks evaluated directly and two novel tasks evaluated after 500 adaptation steps. Task objectives and success rates are reported in Section IV-D and \Cref{tab:real-task-success}. The selected frames illustrate task procedures.}}
	\label{fig:real}
	\vspace*{-0.1cm}

\end{figure*}

\begin{table*}[!t]
\centering
\caption{Detailed Success Rates (\%) on \textcolor{blue}{the Ten ML10} Base Tasks. \textcolor{blue}{The} Tasks \textcolor{blue}{Include Five} Easy, \textcolor{blue}{Three} Medium, and \textcolor{blue}{Two} Hard Tasks.}
\label{tab:main_results_10}
\scriptsize
\setlength{\tabcolsep}{1pt}\renewcommand{\arraystretch}{1.05}
\begin{tabular*}{\textwidth}{@{\extracolsep{\fill}}c cccccccccc}
\toprule
& \multicolumn{5}{c}{Easy} & \multicolumn{3}{c}{Medium} & \multicolumn{2}{c}{Hard} \\
\textcolor{blue}{Method} & \makecell{Door\\Open} & \makecell{Button\\Press\\Topdown} & \makecell{Drawer\\Close} & \makecell{Reach} & \makecell{Window\\Open} & \makecell{Basketball} & \makecell{Peg\\Insert\\Side} & \makecell{Sweep} & \makecell{Pick\\Place} & \makecell{Push} \\
\midrule
\textcolor{blue}{MAML-TRPO} & \mbox{74.00\textcolor{blue}{\,$\pm$\,32.92}} & \mbox{85.00\textcolor{blue}{\,$\pm$\,10.54}} & \mbox{98.33\textcolor{blue}{\,$\pm$\,2.89}} & \mbox{16.00\textcolor{blue}{\,$\pm$\,7.00}} & \mbox{98.67\textcolor{blue}{\,$\pm$\,2.31}} & \mbox{0.33\textcolor{blue}{\,$\pm$\,0.58}} & \mbox{0.00\textcolor{blue}{\,$\pm$\,0.00}} & \mbox{3.67\textcolor{blue}{\,$\pm$\,3.79}} & \mbox{4.00\textcolor{blue}{\,$\pm$\,0.00}} & \mbox{16.00\textcolor{blue}{\,$\pm$\,2.65}} \\
DP3 & \textcolor{blue}{\mbox{82.67\,$\pm$\,2.08}} & \mbox{100.00\textcolor{blue}{\,$\pm$\,0.00}} & \mbox{100.00\textcolor{blue}{\,$\pm$\,0.00}} & \mbox{54.67\textcolor{blue}{\,$\pm$\,4.51}} & \mbox{98.33\textcolor{blue}{\,$\pm$\,2.89}} & \mbox{100.00\textcolor{blue}{\,$\pm$\,0.00}} & \mbox{70.33\textcolor{blue}{\,$\pm$\,10.07}} & \mbox{70.00\textcolor{blue}{\,$\pm$\,6.56}} & \mbox{68.67\textcolor{blue}{\,$\pm$\,5.51}} & \mbox{77.00\textcolor{blue}{\,$\pm$\,3.46}} \\
Mamba & \mbox{86.33\textcolor{blue}{\,$\pm$\,4.04}} & \mbox{100.00\textcolor{blue}{\,$\pm$\,0.00}} & \mbox{100.00\textcolor{blue}{\,$\pm$\,0.00}} & \mbox{39.00\textcolor{blue}{\,$\pm$\,6.93}} & \mbox{100.00\textcolor{blue}{\,$\pm$\,0.00}} & \mbox{100.00\textcolor{blue}{\,$\pm$\,0.00}} & \mbox{62.33\textcolor{blue}{\,$\pm$\,11.85}} & \mbox{66.33\textcolor{blue}{\,$\pm$\,5.51}} & \mbox{0.00\textcolor{blue}{\,$\pm$\,0.00}} & \mbox{67.67\textcolor{blue}{\,$\pm$\,5.03}} \\
FreqPolicy & \mbox{96.67\textcolor{blue}{\,$\pm$\,2.89}} & \mbox{89.00\textcolor{blue}{\,$\pm$\,9.85}} & \mbox{100.00\textcolor{blue}{\,$\pm$\,0.00}} & \mbox{42.33\textcolor{blue}{\,$\pm$\,4.51}} & \mbox{93.67\textcolor{blue}{\,$\pm$\,2.31}} & \mbox{100.00\textcolor{blue}{\,$\pm$\,0.00}} & \mbox{65.33\textcolor{blue}{\,$\pm$\,3.06}} & \mbox{81.00\textcolor{blue}{\,$\pm$\,6.93}} & \mbox{75.00\textcolor{blue}{\,$\pm$\,12.29}} & \mbox{64.67\textcolor{blue}{\,$\pm$\,13.58}} \\
FlowPolicy & \mbox{91.33\textcolor{blue}{\,$\pm$\,1.15}} & \mbox{100.00\textcolor{blue}{\,$\pm$\,0.00}} & \mbox{100.00\textcolor{blue}{\,$\pm$\,0.00}} & \mbox{27.33\textcolor{blue}{\,$\pm$\,2.52}} & \mbox{100.00\textcolor{blue}{\,$\pm$\,0.00}} & \mbox{96.67\textcolor{blue}{\,$\pm$\,3.51}} & \mbox{69.33\textcolor{blue}{\,$\pm$\,11.02}} & \mbox{57.33\textcolor{blue}{\,$\pm$\,4.51}} & \mbox{61.00\textcolor{blue}{\,$\pm$\,9.54}} & \mbox{77.33\textcolor{blue}{\,$\pm$\,2.52}} \\
\textbf{Ours (DAMI)} & \mbox{92.67\textcolor{blue}{\,$\pm$\,2.08}} & \mbox{100.00\textcolor{blue}{\,$\pm$\,0.00}} & \mbox{100.00\textcolor{blue}{\,$\pm$\,0.00}} & \mbox{58.67\textcolor{blue}{\,$\pm$\,5.03}} & \mbox{100.00\textcolor{blue}{\,$\pm$\,0.00}} & \mbox{100.00\textcolor{blue}{\,$\pm$\,0.00}} & \mbox{88.00\textcolor{blue}{\,$\pm$\,8.54}} & \mbox{87.33\textcolor{blue}{\,$\pm$\,3.51}} & \mbox{70.33\textcolor{blue}{\,$\pm$\,2.31}} & \mbox{75.33\textcolor{blue}{\,$\pm$\,12.34}} \\
\bottomrule\end{tabular*}

\par\smallskip\begin{minipage}{\textwidth}\fontsize{7}{8.1}\selectfont\raggedright \textcolor{blue}{Mean $\pm$ SD over three evaluation seeds.}\end{minipage}
\par\vspace{-0.2cm}
\end{table*}

\textbf{Implementation Details.} \textcolor{blue}{The imitation-learning} methods \textcolor{blue}{trained in this work use} a three-layer \textcolor{blue}{visual} MLP\textcolor{blue}{, max pooling}, a projection head, and LayerNorm. We \textcolor{blue}{use a frozen} CLIP ViT-B/32 text encoder~\cite{radford2021learning} \textcolor{blue}{and farthest point sampling} (FPS) \textcolor{blue}{with} 512 points \textcolor{blue}{in} simulation and 1024 \textcolor{blue}{on real robots}. \textcolor{blue}{DAMI trains} for 3000 epochs \textcolor{blue}{with support/query batches} of 20\textcolor{blue}{/}88 \textcolor{blue}{windows} per task \textcolor{blue}{and one inner-loop} gradient update. \textcolor{blue}{Key hyperparameters are listed in \Cref{tab:hyperparams}.}

\textcolor{blue}{In ML10, reference states and point clouds retain their original scales during meta-training and are normalized during adaptation and inference. Reference actions are normalized except at the first meta-training update.}

\textcolor{blue}{Meta-World references contain 200 steps. U2T fuses 204 tokens in a common 512-dimensional space: 200 demonstration tokens, one text token, one observation--timestep token, and two learnable queries. The first query output conditions feature modulation. The 200 positions in \Cref{tab:hyperparams} count demonstration tokens only.}

\textbf{Key Hardware and Software.} \textcolor{blue}{The imitation-learning} methods \textcolor{blue}{trained in this work use} PyTorch on \textcolor{blue}{an} Intel Xeon Platinum 8358 CPU and \textcolor{blue}{an} NVIDIA L40S GPU (48GB).

\begin{table*}[!t]
\centering
\caption{Detailed Success Rates (\%) on All 45 \textcolor{blue}{ML45} Base Tasks. The Tasks \textcolor{blue}{Include} 26 Easy, 9 Medium, 5 Hard, and 5 Very Hard Tasks.}
\label{tab:main_results_45}
\fontsize{6}{7.2}\selectfont
\setlength{\tabcolsep}{0.7pt}\renewcommand{\arraystretch}{1.05}
\begin{tabular*}{\textwidth}{@{\extracolsep{\fill}}c ccccccccccccc}
\toprule
& \multicolumn{13}{c}{Easy, Part I} \\
\textcolor{blue}{Method} & \makecell{Button\\Press\\Topdown} & \makecell{Button\\Press\\Topdown\\Wall} & \makecell{Button\\Press} & \makecell{Button\\Press\\Wall} & \makecell{Coffee\\Button} & \makecell{Dial\\Turn} & \makecell{Door\\Close} & \makecell{Door\\Open} & \makecell{Drawer\\Close} & \makecell{Drawer\\Open} & \makecell{Faucet\\Close} & \makecell{Faucet\\Open} & \makecell{Handle\\Press\\Side} \\
\midrule
\textcolor{blue}{MAML-TRPO} & \mbox{0.00\textcolor{blue}{\,$\pm$\,0.00}} & \mbox{0.00\textcolor{blue}{\,$\pm$\,0.00}} & \mbox{0.00\textcolor{blue}{\,$\pm$\,0.00}} & \mbox{6.00\textcolor{blue}{\,$\pm$\,10.39}} & \mbox{25.33\textcolor{blue}{\,$\pm$\,43.88}} & \mbox{27.33\textcolor{blue}{\,$\pm$\,43.92}} & \mbox{1.33\textcolor{blue}{\,$\pm$\,1.53}} & \mbox{0.00\textcolor{blue}{\,$\pm$\,0.00}} & \mbox{94.67\textcolor{blue}{\,$\pm$\,4.62}} & \mbox{0.00\textcolor{blue}{\,$\pm$\,0.00}} & \mbox{0.00\textcolor{blue}{\,$\pm$\,0.00}} & \mbox{0.67\textcolor{blue}{\,$\pm$\,1.15}} & \mbox{73.33\textcolor{blue}{\,$\pm$\,23.18}} \\
DP3 & \mbox{91.00\textcolor{blue}{\,$\pm$\,4.58}} & \mbox{30.00\textcolor{blue}{\,$\pm$\,10.00}} & \mbox{51.67\textcolor{blue}{\,$\pm$\,6.66}} & \mbox{100.00\textcolor{blue}{\,$\pm$\,0.00}} & \mbox{96.67\textcolor{blue}{\,$\pm$\,4.16}} & \mbox{18.67\textcolor{blue}{\,$\pm$\,4.04}} & \mbox{100.00\textcolor{blue}{\,$\pm$\,0.00}} & \mbox{100.00\textcolor{blue}{\,$\pm$\,0.00}} & \mbox{100.00\textcolor{blue}{\,$\pm$\,0.00}} & \mbox{95.33\textcolor{blue}{\,$\pm$\,4.51}} & \mbox{2.33\textcolor{blue}{\,$\pm$\,2.31}} & \mbox{74.33\textcolor{blue}{\,$\pm$\,2.31}} & \mbox{35.00\textcolor{blue}{\,$\pm$\,1.73}} \\
Mamba & \mbox{100.00\textcolor{blue}{\,$\pm$\,0.00}} & \mbox{73.33\textcolor{blue}{\,$\pm$\,22.55}} & \mbox{50.33\textcolor{blue}{\,$\pm$\,8.08}} & \mbox{100.00\textcolor{blue}{\,$\pm$\,0.00}} & \mbox{92.33\textcolor{blue}{\,$\pm$\,8.08}} & \mbox{22.00\textcolor{blue}{\,$\pm$\,6.93}} & \mbox{100.00\textcolor{blue}{\,$\pm$\,0.00}} & \mbox{99.33\textcolor{blue}{\,$\pm$\,1.15}} & \mbox{100.00\textcolor{blue}{\,$\pm$\,0.00}} & \mbox{98.67\textcolor{blue}{\,$\pm$\,1.53}} & \mbox{4.00\textcolor{blue}{\,$\pm$\,2.65}} & \mbox{100.00\textcolor{blue}{\,$\pm$\,0.00}} & \mbox{55.67\textcolor{blue}{\,$\pm$\,2.08}} \\
FreqPolicy & \mbox{73.33\textcolor{blue}{\,$\pm$\,14.57}} & \mbox{18.33\textcolor{blue}{\,$\pm$\,5.77}} & \mbox{57.67\textcolor{blue}{\,$\pm$\,3.06}} & \mbox{100.00\textcolor{blue}{\,$\pm$\,0.00}} & \mbox{88.00\textcolor{blue}{\,$\pm$\,12.29}} & \mbox{23.33\textcolor{blue}{\,$\pm$\,4.93}} & \mbox{100.00\textcolor{blue}{\,$\pm$\,0.00}} & \mbox{90.33\textcolor{blue}{\,$\pm$\,5.03}} & \mbox{100.00\textcolor{blue}{\,$\pm$\,0.00}} & \mbox{88.00\textcolor{blue}{\,$\pm$\,2.00}} & \mbox{7.33\textcolor{blue}{\,$\pm$\,5.03}} & \mbox{76.00\textcolor{blue}{\,$\pm$\,10.82}} & \mbox{21.33\textcolor{blue}{\,$\pm$\,3.51}} \\
FlowPolicy & \mbox{99.67\textcolor{blue}{\,$\pm$\,0.58}} & \mbox{35.67\textcolor{blue}{\,$\pm$\,5.13}} & \mbox{47.00\textcolor{blue}{\,$\pm$\,9.85}} & \mbox{100.00\textcolor{blue}{\,$\pm$\,0.00}} & \mbox{100.00\textcolor{blue}{\,$\pm$\,0.00}} & \mbox{21.33\textcolor{blue}{\,$\pm$\,2.52}} & \mbox{100.00\textcolor{blue}{\,$\pm$\,0.00}} & \mbox{100.00\textcolor{blue}{\,$\pm$\,0.00}} & \mbox{100.00\textcolor{blue}{\,$\pm$\,0.00}} & \mbox{93.00\textcolor{blue}{\,$\pm$\,2.00}} & \mbox{2.67\textcolor{blue}{\,$\pm$\,3.06}} & \mbox{99.67\textcolor{blue}{\,$\pm$\,0.58}} & \mbox{26.00\textcolor{blue}{\,$\pm$\,3.46}} \\
\textbf{Ours (DAMI)} & \mbox{100.00\textcolor{blue}{\,$\pm$\,0.00}} & \mbox{100.00\textcolor{blue}{\,$\pm$\,0.00}} & \mbox{100.00\textcolor{blue}{\,$\pm$\,0.00}} & \mbox{100.00\textcolor{blue}{\,$\pm$\,0.00}} & \mbox{100.00\textcolor{blue}{\,$\pm$\,0.00}} & \mbox{73.67\textcolor{blue}{\,$\pm$\,7.77}} & \mbox{100.00\textcolor{blue}{\,$\pm$\,0.00}} & \mbox{95.00\textcolor{blue}{\,$\pm$\,0.00}} & \mbox{100.00\textcolor{blue}{\,$\pm$\,0.00}} & \mbox{100.00\textcolor{blue}{\,$\pm$\,0.00}} & \mbox{100.00\textcolor{blue}{\,$\pm$\,0.00}} & \mbox{100.00\textcolor{blue}{\,$\pm$\,0.00}} & \mbox{66.67\textcolor{blue}{\,$\pm$\,7.57}} \\
\bottomrule\end{tabular*}
\par\smallskip
\fontsize{6}{7.2}\selectfont
\setlength{\tabcolsep}{0.7pt}\renewcommand{\arraystretch}{1.05}
\begin{tabular*}{\textwidth}{@{\extracolsep{\fill}}c ccccccccccccc}
\toprule
& \multicolumn{13}{c}{Easy, Part II} \\
\textcolor{blue}{Method} & \makecell{Handle\\Press} & \makecell{Handle\\Pull\\Side} & \makecell{Handle\\Pull} & \makecell{Lever\\Pull} & \makecell{Peg\\Unplug\\Side} & \makecell{Plate\\Slide\\Back\\Side} & \makecell{Plate\\Slide\\Back} & \makecell{Plate\\Slide\\Side} & \makecell{Plate\\Slide} & \makecell{Reach} & \makecell{Reach\\Wall} & \makecell{Window\\Close} & \makecell{Window\\Open} \\
\midrule
\textcolor{blue}{MAML-TRPO} & \mbox{95.33\textcolor{blue}{\,$\pm$\,5.03}} & \mbox{15.33\textcolor{blue}{\,$\pm$\,26.56}} & \mbox{2.00\textcolor{blue}{\,$\pm$\,3.46}} & \mbox{0.00\textcolor{blue}{\,$\pm$\,0.00}} & \mbox{0.67\textcolor{blue}{\,$\pm$\,1.15}} & \mbox{0.00\textcolor{blue}{\,$\pm$\,0.00}} & \mbox{2.00\textcolor{blue}{\,$\pm$\,3.46}} & \mbox{0.00\textcolor{blue}{\,$\pm$\,0.00}} & \mbox{0.00\textcolor{blue}{\,$\pm$\,0.00}} & \mbox{0.00\textcolor{blue}{\,$\pm$\,0.00}} & \mbox{0.00\textcolor{blue}{\,$\pm$\,0.00}} & \mbox{0.00\textcolor{blue}{\,$\pm$\,0.00}} & \mbox{0.00\textcolor{blue}{\,$\pm$\,0.00}} \\
DP3 & \mbox{86.00\textcolor{blue}{\,$\pm$\,7.81}} & \mbox{26.33\textcolor{blue}{\,$\pm$\,12.50}} & \mbox{2.33\textcolor{blue}{\,$\pm$\,3.21}} & \mbox{8.00\textcolor{blue}{\,$\pm$\,3.46}} & \mbox{24.00\textcolor{blue}{\,$\pm$\,3.00}} & \mbox{100.00\textcolor{blue}{\,$\pm$\,0.00}} & \mbox{100.00\textcolor{blue}{\,$\pm$\,0.00}} & \mbox{0.00\textcolor{blue}{\,$\pm$\,0.00}} & \mbox{99.67\textcolor{blue}{\,$\pm$\,0.58}} & \mbox{41.33\textcolor{blue}{\,$\pm$\,9.87}} & \mbox{38.67\textcolor{blue}{\,$\pm$\,9.45}} & \mbox{98.00\textcolor{blue}{\,$\pm$\,2.00}} & \mbox{27.00\textcolor{blue}{\,$\pm$\,5.57}} \\
Mamba & \mbox{92.33\textcolor{blue}{\,$\pm$\,3.21}} & \mbox{22.33\textcolor{blue}{\,$\pm$\,7.57}} & \mbox{5.33\textcolor{blue}{\,$\pm$\,4.16}} & \mbox{16.67\textcolor{blue}{\,$\pm$\,7.64}} & \mbox{30.00\textcolor{blue}{\,$\pm$\,4.36}} & \mbox{100.00\textcolor{blue}{\,$\pm$\,0.00}} & \mbox{100.00\textcolor{blue}{\,$\pm$\,0.00}} & \mbox{0.00\textcolor{blue}{\,$\pm$\,0.00}} & \mbox{98.67\textcolor{blue}{\,$\pm$\,1.15}} & \mbox{33.33\textcolor{blue}{\,$\pm$\,16.80}} & \mbox{41.33\textcolor{blue}{\,$\pm$\,16.80}} & \mbox{100.00\textcolor{blue}{\,$\pm$\,0.00}} & \mbox{33.00\textcolor{blue}{\,$\pm$\,11.36}} \\
FreqPolicy & \mbox{74.00\textcolor{blue}{\,$\pm$\,14.93}} & \mbox{35.00\textcolor{blue}{\,$\pm$\,6.56}} & \mbox{5.33\textcolor{blue}{\,$\pm$\,3.51}} & \mbox{36.67\textcolor{blue}{\,$\pm$\,5.51}} & \mbox{13.67\textcolor{blue}{\,$\pm$\,1.15}} & \mbox{100.00\textcolor{blue}{\,$\pm$\,0.00}} & \mbox{100.00\textcolor{blue}{\,$\pm$\,0.00}} & \mbox{0.00\textcolor{blue}{\,$\pm$\,0.00}} & \mbox{98.67\textcolor{blue}{\,$\pm$\,2.31}} & \mbox{35.00\textcolor{blue}{\,$\pm$\,15.62}} & \mbox{53.67\textcolor{blue}{\,$\pm$\,14.05}} & \mbox{100.00\textcolor{blue}{\,$\pm$\,0.00}} & \mbox{9.33\textcolor{blue}{\,$\pm$\,5.77}} \\
FlowPolicy & \mbox{92.33\textcolor{blue}{\,$\pm$\,1.53}} & \mbox{5.00\textcolor{blue}{\,$\pm$\,1.73}} & \mbox{8.67\textcolor{blue}{\,$\pm$\,2.08}} & \mbox{41.67\textcolor{blue}{\,$\pm$\,5.03}} & \mbox{18.67\textcolor{blue}{\,$\pm$\,2.31}} & \mbox{100.00\textcolor{blue}{\,$\pm$\,0.00}} & \mbox{97.33\textcolor{blue}{\,$\pm$\,1.53}} & \mbox{0.00\textcolor{blue}{\,$\pm$\,0.00}} & \mbox{92.00\textcolor{blue}{\,$\pm$\,3.00}} & \mbox{39.67\textcolor{blue}{\,$\pm$\,14.98}} & \mbox{40.67\textcolor{blue}{\,$\pm$\,4.51}} & \mbox{100.00\textcolor{blue}{\,$\pm$\,0.00}} & \mbox{30.33\textcolor{blue}{\,$\pm$\,9.07}} \\
\textbf{Ours (DAMI)} & \mbox{86.67\textcolor{blue}{\,$\pm$\,0.58}} & \mbox{37.33\textcolor{blue}{\,$\pm$\,3.79}} & \mbox{16.00\textcolor{blue}{\,$\pm$\,4.36}} & \mbox{75.00\textcolor{blue}{\,$\pm$\,3.00}} & \mbox{89.00\textcolor{blue}{\,$\pm$\,3.00}} & \mbox{100.00\textcolor{blue}{\,$\pm$\,0.00}} & \mbox{100.00\textcolor{blue}{\,$\pm$\,0.00}} & \mbox{99.33\textcolor{blue}{\,$\pm$\,1.15}} & \mbox{98.33\textcolor{blue}{\,$\pm$\,2.89}} & \mbox{63.00\textcolor{blue}{\,$\pm$\,1.00}} & \mbox{68.67\textcolor{blue}{\,$\pm$\,18.77}} & \mbox{100.00\textcolor{blue}{\,$\pm$\,0.00}} & \mbox{98.33\textcolor{blue}{\,$\pm$\,2.89}} \\
\bottomrule\end{tabular*}
\par\smallskip
\fontsize{6}{7.2}\selectfont
\setlength{\tabcolsep}{0.7pt}\renewcommand{\arraystretch}{1.05}
\begin{tabular*}{\textwidth}{@{\extracolsep{\fill}}c ccccccccc}
\toprule
& \multicolumn{9}{c}{Medium} \\
\textcolor{blue}{Method} & \makecell{Basketball} & \makecell{Coffee\\Pull} & \makecell{Coffee\\Push} & \makecell{Hammer} & \makecell{Peg\\Insert\\Side} & \makecell{Push\\Wall} & \makecell{Soccer} & \makecell{Sweep\\Into} & \makecell{Sweep} \\
\midrule
\textcolor{blue}{MAML-TRPO} & \mbox{0.00\textcolor{blue}{\,$\pm$\,0.00}} & \mbox{0.00\textcolor{blue}{\,$\pm$\,0.00}} & \mbox{8.00\textcolor{blue}{\,$\pm$\,8.00}} & \mbox{0.00\textcolor{blue}{\,$\pm$\,0.00}} & \mbox{0.00\textcolor{blue}{\,$\pm$\,0.00}} & \mbox{0.00\textcolor{blue}{\,$\pm$\,0.00}} & \mbox{3.33\textcolor{blue}{\,$\pm$\,4.16}} & \mbox{1.33\textcolor{blue}{\,$\pm$\,2.31}} & \mbox{0.00\textcolor{blue}{\,$\pm$\,0.00}} \\
DP3 & \mbox{96.00\textcolor{blue}{\,$\pm$\,4.58}} & \mbox{0.00\textcolor{blue}{\,$\pm$\,0.00}} & \mbox{65.67\textcolor{blue}{\,$\pm$\,15.53}} & \mbox{30.00\textcolor{blue}{\,$\pm$\,6.56}} & \mbox{2.33\textcolor{blue}{\,$\pm$\,2.52}} & \mbox{47.67\textcolor{blue}{\,$\pm$\,4.73}} & \mbox{18.33\textcolor{blue}{\,$\pm$\,5.69}} & \mbox{15.00\textcolor{blue}{\,$\pm$\,5.00}} & \mbox{39.33\textcolor{blue}{\,$\pm$\,10.02}} \\
Mamba & \mbox{100.00\textcolor{blue}{\,$\pm$\,0.00}} & \mbox{0.00\textcolor{blue}{\,$\pm$\,0.00}} & \mbox{58.67\textcolor{blue}{\,$\pm$\,10.07}} & \mbox{27.67\textcolor{blue}{\,$\pm$\,3.21}} & \mbox{3.00\textcolor{blue}{\,$\pm$\,2.65}} & \mbox{56.67\textcolor{blue}{\,$\pm$\,12.42}} & \mbox{13.00\textcolor{blue}{\,$\pm$\,6.08}} & \mbox{23.33\textcolor{blue}{\,$\pm$\,7.23}} & \mbox{46.00\textcolor{blue}{\,$\pm$\,4.58}} \\
FreqPolicy & \mbox{95.33\textcolor{blue}{\,$\pm$\,1.53}} & \mbox{0.00\textcolor{blue}{\,$\pm$\,0.00}} & \mbox{68.67\textcolor{blue}{\,$\pm$\,9.71}} & \mbox{16.00\textcolor{blue}{\,$\pm$\,2.00}} & \mbox{5.33\textcolor{blue}{\,$\pm$\,0.58}} & \mbox{41.00\textcolor{blue}{\,$\pm$\,23.07}} & \mbox{13.33\textcolor{blue}{\,$\pm$\,2.08}} & \mbox{18.33\textcolor{blue}{\,$\pm$\,4.51}} & \mbox{35.67\textcolor{blue}{\,$\pm$\,10.02}} \\
FlowPolicy & \mbox{69.00\textcolor{blue}{\,$\pm$\,4.36}} & \mbox{0.00\textcolor{blue}{\,$\pm$\,0.00}} & \mbox{59.33\textcolor{blue}{\,$\pm$\,5.51}} & \mbox{15.33\textcolor{blue}{\,$\pm$\,8.02}} & \mbox{2.67\textcolor{blue}{\,$\pm$\,3.06}} & \mbox{43.00\textcolor{blue}{\,$\pm$\,9.17}} & \mbox{7.33\textcolor{blue}{\,$\pm$\,2.52}} & \mbox{17.33\textcolor{blue}{\,$\pm$\,9.50}} & \mbox{45.33\textcolor{blue}{\,$\pm$\,7.37}} \\
\textbf{Ours (DAMI)} & \mbox{97.33\textcolor{blue}{\,$\pm$\,2.08}} & \mbox{82.33\textcolor{blue}{\,$\pm$\,8.50}} & \mbox{92.67\textcolor{blue}{\,$\pm$\,2.31}} & \mbox{92.00\textcolor{blue}{\,$\pm$\,2.00}} & \mbox{59.67\textcolor{blue}{\,$\pm$\,7.57}} & \mbox{80.00\textcolor{blue}{\,$\pm$\,12.12}} & \mbox{28.33\textcolor{blue}{\,$\pm$\,5.51}} & \mbox{15.00\textcolor{blue}{\,$\pm$\,7.00}} & \mbox{92.00\textcolor{blue}{\,$\pm$\,1.00}} \\
\bottomrule\end{tabular*}
\par\smallskip
\fontsize{6}{7.2}\selectfont
\setlength{\tabcolsep}{0.7pt}\renewcommand{\arraystretch}{1.05}
\begin{tabular*}{\textwidth}{@{\extracolsep{\fill}}c ccccccccccc}
\toprule
& \multicolumn{11}{c}{Hard and Very Hard} \\
\textcolor{blue}{Method} & \makecell{Assembly} & \makecell{Pick\\Out\\Of\\Hole} & \makecell{Pick\\Place} & \makecell{Push\\Back} & \makecell{Push} & \makecell{Disassemble} & \makecell{Pick\\Place\\Wall} & \makecell{Shelf\\Place} & \makecell{Stick\\Pull} & \makecell{Stick\\Push} & \makecell{Avg.} \\
\midrule
\textcolor{blue}{MAML-TRPO} & \mbox{0.00\textcolor{blue}{\,$\pm$\,0.00}} & \mbox{0.00\textcolor{blue}{\,$\pm$\,0.00}} & \mbox{0.00\textcolor{blue}{\,$\pm$\,0.00}} & \mbox{0.00\textcolor{blue}{\,$\pm$\,0.00}} & \mbox{0.00\textcolor{blue}{\,$\pm$\,0.00}} & \mbox{0.00\textcolor{blue}{\,$\pm$\,0.00}} & \mbox{0.00\textcolor{blue}{\,$\pm$\,0.00}} & \mbox{0.00\textcolor{blue}{\,$\pm$\,0.00}} & \mbox{0.00\textcolor{blue}{\,$\pm$\,0.00}} & \mbox{0.00\textcolor{blue}{\,$\pm$\,0.00}} & \mbox{7.93\textcolor{blue}{\,$\pm$\,1.82}} \\
DP3 & \mbox{34.67\textcolor{blue}{\,$\pm$\,13.61}} & \mbox{28.67\textcolor{blue}{\,$\pm$\,12.22}} & \mbox{0.00\textcolor{blue}{\,$\pm$\,0.00}} & \mbox{46.67\textcolor{blue}{\,$\pm$\,8.33}} & \mbox{8.00\textcolor{blue}{\,$\pm$\,3.61}} & \mbox{28.00\textcolor{blue}{\,$\pm$\,10.54}} & \mbox{22.00\textcolor{blue}{\,$\pm$\,4.00}} & \mbox{5.67\textcolor{blue}{\,$\pm$\,6.43}} & \mbox{16.67\textcolor{blue}{\,$\pm$\,6.66}} & \mbox{74.67\textcolor{blue}{\,$\pm$\,8.08}} & \mbox{47.24\textcolor{blue}{\,$\pm$\,1.01}} \\
Mamba & \mbox{39.00\textcolor{blue}{\,$\pm$\,15.72}} & \mbox{36.33\textcolor{blue}{\,$\pm$\,14.19}} & \mbox{36.33\textcolor{blue}{\,$\pm$\,14.19}} & \mbox{46.33\textcolor{blue}{\,$\pm$\,4.04}} & \mbox{6.00\textcolor{blue}{\,$\pm$\,1.00}} & \mbox{4.67\textcolor{blue}{\,$\pm$\,4.16}} & \mbox{21.67\textcolor{blue}{\,$\pm$\,5.86}} & \mbox{4.67\textcolor{blue}{\,$\pm$\,2.52}} & \mbox{28.33\textcolor{blue}{\,$\pm$\,5.69}} & \mbox{70.00\textcolor{blue}{\,$\pm$\,10.54}} & \mbox{50.90\textcolor{blue}{\,$\pm$\,0.84}} \\
FreqPolicy & \mbox{42.67\textcolor{blue}{\,$\pm$\,13.05}} & \mbox{40.00\textcolor{blue}{\,$\pm$\,3.46}} & \mbox{2.67\textcolor{blue}{\,$\pm$\,2.52}} & \mbox{38.33\textcolor{blue}{\,$\pm$\,12.66}} & \mbox{6.33\textcolor{blue}{\,$\pm$\,0.58}} & \mbox{14.33\textcolor{blue}{\,$\pm$\,2.52}} & \mbox{27.00\textcolor{blue}{\,$\pm$\,15.72}} & \mbox{3.67\textcolor{blue}{\,$\pm$\,5.51}} & \mbox{13.67\textcolor{blue}{\,$\pm$\,8.14}} & \mbox{76.67\textcolor{blue}{\,$\pm$\,5.51}} & \mbox{45.87\textcolor{blue}{\,$\pm$\,1.40}} \\
FlowPolicy & \mbox{24.00\textcolor{blue}{\,$\pm$\,9.85}} & \mbox{8.33\textcolor{blue}{\,$\pm$\,10.12}} & \mbox{0.00\textcolor{blue}{\,$\pm$\,0.00}} & \mbox{43.00\textcolor{blue}{\,$\pm$\,8.00}} & \mbox{3.67\textcolor{blue}{\,$\pm$\,3.21}} & \mbox{14.00\textcolor{blue}{\,$\pm$\,6.24}} & \mbox{20.00\textcolor{blue}{\,$\pm$\,2.00}} & \mbox{10.67\textcolor{blue}{\,$\pm$\,3.79}} & \mbox{16.33\textcolor{blue}{\,$\pm$\,3.51}} & \mbox{72.67\textcolor{blue}{\,$\pm$\,8.33}} & \mbox{45.85\textcolor{blue}{\,$\pm$\,1.01}} \\
\textbf{Ours (DAMI)} & \mbox{88.00\textcolor{blue}{\,$\pm$\,9.85}} & \mbox{31.67\textcolor{blue}{\,$\pm$\,8.96}} & \mbox{60.33\textcolor{blue}{\,$\pm$\,4.04}} & \mbox{44.67\textcolor{blue}{\,$\pm$\,3.79}} & \mbox{83.33\textcolor{blue}{\,$\pm$\,3.21}} & \mbox{65.00\textcolor{blue}{\,$\pm$\,10.15}} & \mbox{86.33\textcolor{blue}{\,$\pm$\,0.58}} & \mbox{23.67\textcolor{blue}{\,$\pm$\,8.08}} & \mbox{69.33\textcolor{blue}{\,$\pm$\,15.01}} & \mbox{98.67\textcolor{blue}{\,$\pm$\,1.15}} & \mbox{79.05\textcolor{blue}{\,$\pm$\,1.24}} \\
\bottomrule\end{tabular*}
\par\smallskip

\par\smallskip\begin{minipage}{\textwidth}\fontsize{7}{8.1}\selectfont\raggedright \textcolor{blue}{Mean $\pm$ SD over three evaluation seeds.}\end{minipage}
\end{table*}

\subsection{Quantitative Comparison}
We \textcolor{blue}{evaluate} DAMI on Meta-World \textcolor{blue}{ML10/ML45} and RLBench \textcolor{blue}{FS25}. Following DP3\textcolor{blue}{~}\cite{Ze2024DP3}, Meta-World tasks \textcolor{blue}{are grouped as} Easy (E), Medium (M), Hard (H), and Very Hard (VH). \textcolor{blue}{Parenthesized numbers give the task count per group. Where reported, $\pm$ denotes sample SD across evaluation seeds, with single-seed results identified separately. Bold and underlined entries mark numerical rankings.}

\subsubsection{Novel Tasks Evaluation}
\label{sim_eval}
\textcolor{blue}{After few-shot fine-tuning, DAMI achieves mean success rates of 56.80\%, 37.53\%, and 10.80\% on ML10, ML45, and RLBench FS25, respectively (\Cref{unseen-result-table,tab:rlbench_fs25}). The strongest baseline in each setting achieves 48.87\%, 37.53\%, and 9.93\%, placing DAMI first on ML10 and RLBench FS25 and jointly first on ML45. On RLBench FS25, DAMI records successful rollouts on \textit{Open Door} and \textit{Open Oven} (\Cref{tab:rlbench_fs25}).}

\subsubsection{Base Tasks Evaluation}
\textcolor{blue}{On} base tasks, \Cref{seen-result-table} summarizes results by difficulty, \textcolor{blue}{and \Cref{tab:main_results_10,tab:main_results_45} report the} 10 ML10 and 45 ML45 \textcolor{blue}{task results}. DAMI achieves average success rates of 87.23\% on ML10 and 79.05\% on ML45\textcolor{blue}{, the highest means among the compared methods}.

\begin{table*}[!t]
\centering
\caption{\textcolor{blue}{Comparison with MAML-Enhanced Baselines} on Meta-World ML10. Success Rates (\%) \textcolor{blue}{Are Reported for} Base-Task Categories and Individual Novel Tasks.}
\label{tab:ml10_maml_summary}
\fontsize{6.5}{7.8}\selectfont
\setlength{\tabcolsep}{1pt}\renewcommand{\arraystretch}{1.05}
\begin{tabular*}{\textwidth}{@{\extracolsep{\fill}}c cccccccccc}
\toprule
& \multicolumn{4}{c}{Base Tasks} & \multicolumn{6}{c}{Novel Tasks} \\ \cmidrule(lr){2-5}\cmidrule(lr){6-11}
\textcolor{blue}{Method} & E(5) & M(3) & H(2) & Avg. & \makecell{Door\\Close} & \makecell{Drawer\\Open} & \makecell{Lever\\Pull} & \makecell{Sweep\\Into} & \makecell{Shelf\\Place} & Avg. \\
\midrule
\textcolor{blue}{MAML-DP3} & \mbox{86.13\textcolor{blue}{\,$\pm$\,0.64}} & \mbox{\underline{80.22}\textcolor{blue}{\,$\pm$\,5.32}} & \mbox{\textbf{73.00}\textcolor{blue}{\,$\pm$\,8.67}} & \mbox{\underline{81.73}\textcolor{blue}{\,$\pm$\,0.90}} & \mbox{\textbf{100.00}\textcolor{blue}{\,$\pm$\,0.00}} & \mbox{\textbf{100.00}\textcolor{blue}{\,$\pm$\,0.00}} & \mbox{\underline{46.67}\textcolor{blue}{\,$\pm$\,27.54}} & \mbox{\underline{20.33}\textcolor{blue}{\,$\pm$\,4.73}} & \mbox{\underline{8.33}\textcolor{blue}{\,$\pm$\,3.21}} & \mbox{\underline{55.07}\textcolor{blue}{\,$\pm$\,6.70}} \\
\textcolor{blue}{MAML-Mamba} & \mbox{\underline{87.33}\textcolor{blue}{\,$\pm$\,1.50}} & \mbox{79.11\textcolor{blue}{\,$\pm$\,2.80}} & \mbox{33.33\textcolor{blue}{\,$\pm$\,4.01}} & \mbox{74.07\textcolor{blue}{\,$\pm$\,0.81}} & \mbox{\textbf{100.00}\textcolor{blue}{\,$\pm$\,0.00}} & \mbox{\textbf{100.00}\textcolor{blue}{\,$\pm$\,0.00}} & \mbox{\textbf{55.33}\textcolor{blue}{\,$\pm$\,6.35}} & \mbox{12.67\textcolor{blue}{\,$\pm$\,4.62}} & \mbox{3.67\textcolor{blue}{\,$\pm$\,1.53}} & \mbox{54.33\textcolor{blue}{\,$\pm$\,0.42}} \\
\textcolor{blue}{MAML-FreqPolicy} & \mbox{85.20\textcolor{blue}{\,$\pm$\,3.60}} & \mbox{78.22\textcolor{blue}{\,$\pm$\,3.47}} & \mbox{70.33\textcolor{blue}{\,$\pm$\,4.65}} & \mbox{80.13\textcolor{blue}{\,$\pm$\,2.81}} & \mbox{\textbf{100.00}\textcolor{blue}{\,$\pm$\,0.00}} & \mbox{\underline{99.33}\textcolor{blue}{\,$\pm$\,1.15}} & \mbox{35.67\textcolor{blue}{\,$\pm$\,9.61}} & \mbox{18.67\textcolor{blue}{\,$\pm$\,11.50}} & \mbox{3.67\textcolor{blue}{\,$\pm$\,3.51}} & \mbox{51.47\textcolor{blue}{\,$\pm$\,0.50}} \\
\textcolor{blue}{MAML-FlowPolicy} & \mbox{84.53\textcolor{blue}{\,$\pm$\,0.50}} & \mbox{77.56\textcolor{blue}{\,$\pm$\,2.22}} & \mbox{62.17\textcolor{blue}{\,$\pm$\,9.28}} & \mbox{77.97\textcolor{blue}{\,$\pm$\,2.14}} & \mbox{\underline{97.00}\textcolor{blue}{\,$\pm$\,5.20}} & \mbox{55.00\textcolor{blue}{\,$\pm$\,19.08}} & \mbox{32.67\textcolor{blue}{\,$\pm$\,23.46}} & \mbox{12.00\textcolor{blue}{\,$\pm$\,1.73}} & \mbox{0.00\textcolor{blue}{\,$\pm$\,0.00}} & \mbox{39.33\textcolor{blue}{\,$\pm$\,0.76}} \\
\rowcolor{gray!15}\textbf{Ours (DAMI)} & \mbox{\textbf{90.27}\textcolor{blue}{\,$\pm$\,1.42}} & \mbox{\textbf{91.78}\textcolor{blue}{\,$\pm$\,4.00}} & \mbox{\underline{72.83}\textcolor{blue}{\,$\pm$\,5.39}} & \mbox{\textbf{87.23}\textcolor{blue}{\,$\pm$\,2.02}} & \mbox{\textbf{100.00}\textcolor{blue}{\,$\pm$\,0.00}} & \mbox{\textbf{100.00}\textcolor{blue}{\,$\pm$\,0.00}} & \mbox{45.33\textcolor{blue}{\,$\pm$\,18.45}} & \mbox{\textbf{28.00}\textcolor{blue}{\,$\pm$\,14.73}} & \mbox{\textbf{10.67}\textcolor{blue}{\,$\pm$\,1.15}} & \mbox{\textbf{56.80}\textcolor{blue}{\,$\pm$\,2.40}} \\
\bottomrule\end{tabular*}

\par\smallskip\begin{minipage}{\textwidth}\fontsize{7}{8.1}\selectfont\raggedright \textcolor{blue}{Mean $\pm$ SD over three evaluation seeds. Bold and underline indicate first and second ranks, including ties. Baselines use MAML.}\end{minipage}

\end{table*}

\subsubsection{\textcolor{blue}{Comparison with MAML-Enhanced Baselines on ML10}}
\textcolor{blue}{\Cref{tab:ml10_maml_summary} compares the original-protocol DAMI model with MAML-enhanced versions of DP3, Mamba, FreqPolicy, and FlowPolicy on ML10.} \textcolor{blue}{DAMI has the highest mean in both task groups, reaching $(87.23 \pm 2.02)\%$ on base tasks and $(56.80 \pm 2.40)\%$ on novel tasks. MAML-DP3 is the strongest baseline in both groups, achieving $(81.73 \pm 0.90)\%$ and $(55.07 \pm 6.70)\%$, respectively.}

\begin{table*}[!t]
\centering
\caption{\textcolor{blue}{ML10 Task Performance under Episode-Disjoint Meta-Training. Success Rates (\%) Are Reported as Mean $\pm$ Sample SD.}}
\label{tab:episode_disjoint}
\begingroup\color{blue}\footnotesize
\setlength{\tabcolsep}{3pt}
\begin{tabular*}{\textwidth}{@{\extracolsep{\fill}}cccccc}
\toprule
\multicolumn{4}{c}{\textit{Base tasks, direct inference}} & \multicolumn{2}{c}{\textit{Novel tasks, 200-step adaptation}} \\
\cmidrule(lr){1-4}\cmidrule(lr){5-6}
Task & Mean $\pm$ SD & Task & Mean $\pm$ SD & Task & Mean $\pm$ SD \\
\midrule
Peg Insert Side & $77.00 \pm 10.15$ & Pick Place & $73.67 \pm 1.53$ & Door Close & $100.00 \pm 0.00$ \\
Reach & $85.67 \pm 2.08$ & Button Press Topdown & $100.00 \pm 0.00$ & Drawer Open & $99.33 \pm 1.15$ \\
Door Open & $92.33 \pm 9.29$ & Push & $65.00 \pm 2.00$ & Lever Pull & $61.33 \pm 20.13$ \\
Basketball & $100.00 \pm 0.00$ & Drawer Close & $100.00 \pm 0.00$ & Sweep Into & $24.33 \pm 3.21$ \\
Window Open & $100.00 \pm 0.00$ & Sweep & $83.67 \pm 7.23$ & Shelf Place & $4.67 \pm 3.06$ \\
\midrule
\multicolumn{3}{c}{\textbf{Base-task average}} & \textbf{$87.73 \pm 2.95$} & \textbf{Novel-task average} & \textbf{$57.93 \pm 4.31$} \\
\bottomrule
\end{tabular*}
\endgroup
\end{table*}

\subsubsection{\textcolor{blue}{Episode-Disjoint Meta-Training on ML10}}
\textcolor{blue}{\Cref{tab:episode_disjoint} reports full DAMI under episode-disjoint meta-training. Its base- and novel-task means are 0.50 and 1.13 percentage points higher than the original-protocol results in \Cref{seen-result-table,unseen-result-table}.}

\begin{table*}[!t]
\centering
\caption{\textcolor{blue}{Input and Fusion Controls on ML10. ``Demo'' Is a Complete Reference Demonstration. Success Rates (\%) Are Means $\pm$ Sample SDs.}}
\label{tab:ml10_matched_conditioning}
\begingroup\color{blue}\footnotesize
\setlength{\tabcolsep}{8pt}
\begin{tabular*}{\textwidth}{@{\extracolsep{\fill}}cccc}
\toprule
Method & Training & Base tasks & Novel tasks \\
\midrule
DP3 + language & BC & $87.67 \pm 0.21$ & $55.07 \pm 2.44$ \\
DP3 + demonstration & BC & $47.77 \pm 4.83$ & $53.40 \pm 5.24$ \\
DP3 + language + demo & BC & $67.67 \pm 0.60$ & $53.13 \pm 5.14$ \\
MAML-DP3 + language + demo & MAML & $70.97 \pm 1.37$ & $53.40 \pm 4.01$ \\
VMT + simple fusion & MAML & $78.23 \pm 2.89$ & $55.20 \pm 3.36$ \\
Full DAMI & MAML & $87.73 \pm 2.95$ & $57.93 \pm 4.31$ \\
\midrule
\multicolumn{2}{c}{Novel-task details} & VMT + simple fusion & Full DAMI \\
\midrule
\multicolumn{2}{c}{Door Close} & $100.00 \pm 0.00$ & $100.00 \pm 0.00$ \\
\multicolumn{2}{c}{Drawer Open} & $97.00 \pm 3.00$ & $99.33 \pm 1.15$ \\
\multicolumn{2}{c}{Lever Pull} & $49.67 \pm 8.50$ & $61.33 \pm 20.13$ \\
\multicolumn{2}{c}{Sweep Into} & $24.33 \pm 6.81$ & $24.33 \pm 3.21$ \\
\multicolumn{2}{c}{Shelf Place} & $5.00 \pm 1.73$ & $4.67 \pm 3.06$ \\
\bottomrule
\end{tabular*}
\endgroup
\end{table*}

\subsubsection{\textcolor{blue}{Conditioning Inputs and Fusion on ML10}}
\textcolor{blue}{\Cref{tab:ml10_matched_conditioning} separates conditioning-input effects from training and fusion design. Language-only DP3 achieves a base-task mean close to DAMI (87.67\% versus 87.73\%) but a lower novel-task mean (55.07\% versus 57.93\%). Adding reference demonstrations to the simple language-conditioned BC model lowers both means, indicating that additional input information alone does not guarantee improvement.}

\textcolor{blue}{With language and reference inputs matched, DAMI exceeds the DP3 and MAML-DP3 controls by 4.80 and 4.53 percentage points on novel tasks. The latter comparison also matches MAML training, supporting DAMI's representation and conditioning design beyond additional input information. With VMT retained, DAMI exceeds the variant using simple fusion by 2.73 points on novel tasks, supporting the combined U2T and TCFM design.}

\subsection{Qualitative Comparison}
\textcolor{blue}{\Cref{fig:qualitative} presents selected executions on base tasks and on novel tasks after task-specific fine-tuning. In the base-task \textit{Door Open} example, DP3 loses contact with the handle, whereas DAMI completes the task. In the novel-task examples, DP3 fails on \textit{Sweep Into} and produces a motion resembling a door-opening interaction when evaluated on \textit{Drawer Open}. DAMI completes the executions shown. Success and failure are determined by the task-success criterion over the complete rollout.}

\textcolor{blue}{These examples illustrate DAMI's ability to maintain effective object interactions and execute task-appropriate motions. Together with the success-rate results, they support its effectiveness on base tasks and its adaptability to novel tasks.}

\begin{figure*}[t]
	\centering
	\includegraphics[width=\textwidth]{picture/real_config.pdf}
	\caption{\textcolor{blue}{\textbf{Initial Configurations for Real-World Evaluation.} Each row corresponds to a task, and Config 1--10 show its ten trial initializations. DAMI, DP3, and MAML-DP3 use the same initialization sequence. The images document starting states, with task outcomes reported separately in \Cref{tab:real-task-success}.}}
	\label{fig:real_config}
	\vspace{-0.25cm}
\end{figure*}

\subsection{Real-world Experiments}
We conduct experiments on a UR10e manipulator equipped with an AG95 gripper and a RealSense D435i camera, adopting the DP3 visual format \cite{Ze2024DP3}. We design seven tasks \textcolor{blue}{comprising five base tasks and two novel tasks.} \textcolor{blue}{We collect} 10 demonstrations \textcolor{blue}{per task} via teleoperation. The real-world experimental setup and representative executions of the designed tasks are shown in \Cref{fig:real}.

The five base tasks are defined as follows:
\begin{itemize}
	\setlength{\itemsep}{3pt}
	\setlength{\parsep}{0pt}
	\setlength{\topsep}{3pt}
	\setlength{\partopsep}{0pt}
	\item \textbf{\textit{Close Drawer}:} The robot closes the top and middle drawers of a three-tier drawer unit.
	\item \textbf{\textit{Object Place}:} The robot grasps a plastic fruit and deposits it into a basket, with the initial fruit position distributed uniformly across experiments.
	\item \textbf{\textit{Press Button}:} The robot presses a tabletop button whose successful activation produces an auditory signal.
	\item \textbf{\textit{\textcolor{blue}{Push} Ball}:} The robot moves a ball into a designated goal \textcolor{blue}{region.}
	\item \textbf{\textit{Sweep Cylinder}:} The robot pushes a cylinder past a yellow marker line. Training uses only a standard green cylinder, whereas testing introduces three types of irregular cylinders to assess \textcolor{blue}{object-level} generalization \textcolor{blue}{within this base task.}
\end{itemize}

The two novel tasks are defined as follows:
\begin{itemize}
	\setlength{\itemsep}{3pt}
	\setlength{\parsep}{0pt}
	\setlength{\topsep}{3pt}
	\setlength{\partopsep}{0pt}
	\item \textbf{\textit{Push Cube}:} A small cube rests on a larger cube and must be pushed off. The adaptation data contain only a purple cube, whereas testing additionally introduces a smaller $2 \times 2$ Rubik's cube.
	\item \textbf{\textit{Stack Cube}:} The robot picks up a small cube from the table and stacks it onto a larger cube.
\end{itemize}

The initial test configurations for all seven tasks are shown in \Cref{fig:real_config}. \textcolor{blue}{For each task, object positions are uniformly distributed over the evaluation region. DAMI, DP3, and MAML-DP3 use the same position distribution and the same sequence of initial configurations.} \textcolor{blue}{Supplementary} video recordings document executions of \textcolor{blue}{all three} methods on \textcolor{blue}{the} evaluation tasks.

\textcolor{blue}{All methods use their last saved training checkpoint. Base tasks use direct inference, whereas each novel task is initialized separately and fine-tuned for 500 steps using its demonstrations. DAMI retains language and reference-demonstration conditioning, and each fixed task policy is evaluated over ten trials. Two-sided 95\% Wilson intervals quantify task-level uncertainty under an independent-trial binomial model.}

\begin{table}[!t]
\centering
\caption{Real-World Task Success Rates (\%). \textcolor{blue}{Brackets Give 95\% Wilson Confidence Intervals. All Methods Use Their Last Saved Training Checkpoint.}}
\label{tab:real-task-success}
\footnotesize
\setlength{\tabcolsep}{1pt}\renewcommand{\arraystretch}{1.05}
\begin{tabular*}{\columnwidth}{@{\extracolsep{\fill}}ccccc}
\toprule
Type & Task & DP3 & \textcolor{blue}{MAML-DP3} & DAMI (Ours) \\
\midrule
Base & Close Drawer & \makecell{90.00 \textcolor{blue}{(9/10)} \\ \textcolor{blue}{[59.58, 98.21]}} & \makecell{\textcolor{blue}{90.00 (9/10)} \\ \textcolor{blue}{[59.58, 98.21]}} & \makecell{\textcolor{blue}{100.00 (10/10)} \\ \textcolor{blue}{[72.25, 100.00]}} \\
 & Object Place & \makecell{\textcolor{blue}{80.00 (8/10)} \\ \textcolor{blue}{[49.02, 94.33]}} & \makecell{\textcolor{blue}{90.00 (9/10)} \\ \textcolor{blue}{[59.58, 98.21]}} & \makecell{\textcolor{blue}{90.00 (9/10)} \\ \textcolor{blue}{[59.58, 98.21]}} \\
 & Press Button & \makecell{\textcolor{blue}{40.00 (4/10)} \\ \textcolor{blue}{[16.82, 68.73]}} & \makecell{\textcolor{blue}{60.00 (6/10)} \\ \textcolor{blue}{[31.27, 83.18]}} & \makecell{\textcolor{blue}{100.00 (10/10)} \\ \textcolor{blue}{[72.25, 100.00]}} \\
 & \textcolor{blue}{Push} Ball & \makecell{100.00 \textcolor{blue}{(10/10)} \\ \textcolor{blue}{[72.25, 100.00]}} & \makecell{\textcolor{blue}{100.00 (10/10)} \\ \textcolor{blue}{[72.25, 100.00]}} & \makecell{100.00 \textcolor{blue}{(10/10)} \\ \textcolor{blue}{[72.25, 100.00]}} \\
 & Sweep Cylinder & \makecell{\textcolor{blue}{80.00 (8/10)} \\ \textcolor{blue}{[49.02, 94.33]}} & \makecell{\textcolor{blue}{90.00 (9/10)} \\ \textcolor{blue}{[59.58, 98.21]}} & \makecell{100.00 \textcolor{blue}{(10/10)} \\ \textcolor{blue}{[72.25, 100.00]}} \\
\midrule
Novel & Push Cube & \makecell{\textcolor{blue}{80.00 (8/10)} \\ \textcolor{blue}{[49.02, 94.33]}} & \makecell{\textcolor{blue}{90.00 (9/10)} \\ \textcolor{blue}{[59.58, 98.21]}} & \makecell{\textcolor{blue}{100.00 (10/10)} \\ \textcolor{blue}{[72.25, 100.00]}} \\
 & Stack Cube & \makecell{\textcolor{blue}{30.00 (3/10)} \\ \textcolor{blue}{[10.78, 60.32]}} & \makecell{\textcolor{blue}{10.00 (1/10)} \\ \textcolor{blue}{[1.79, 40.42]}} & \makecell{\textcolor{blue}{90.00 (9/10)} \\ \textcolor{blue}{[59.58, 98.21]}} \\
\midrule\multicolumn{2}{c}{Average} & \textcolor{blue}{71.43} & \textcolor{blue}{75.71} & \textcolor{blue}{\textbf{97.14}} \\
\bottomrule\end{tabular*}

\par\smallskip\begin{minipage}{\columnwidth}\fontsize{7}{8.1}\selectfont\raggedright \textcolor{blue}{Rates use 10 matched trials per task. Parentheses give successful trials/total trials, and brackets give individual two-sided 95\% Wilson intervals in percent.}\end{minipage}
\par\vspace{-0.25cm}
\end{table}

\textbf{Quantitative Results.}
\textcolor{blue}{DAMI achieves 97.14\% success (68/70 trials), compared with 75.71\% for MAML-DP3 and 71.43\% for DP3 (\Cref{tab:real-task-success}). On the novel tasks, DAMI reaches 100\% on \textit{Push Cube} and 90\% on \textit{Stack Cube}. Corresponding rates are 90\% and 10\% for MAML-DP3, and 80\% and 30\% for DP3.}

\begin{figure}[t]
	\centering
	\includegraphics[width=\columnwidth]{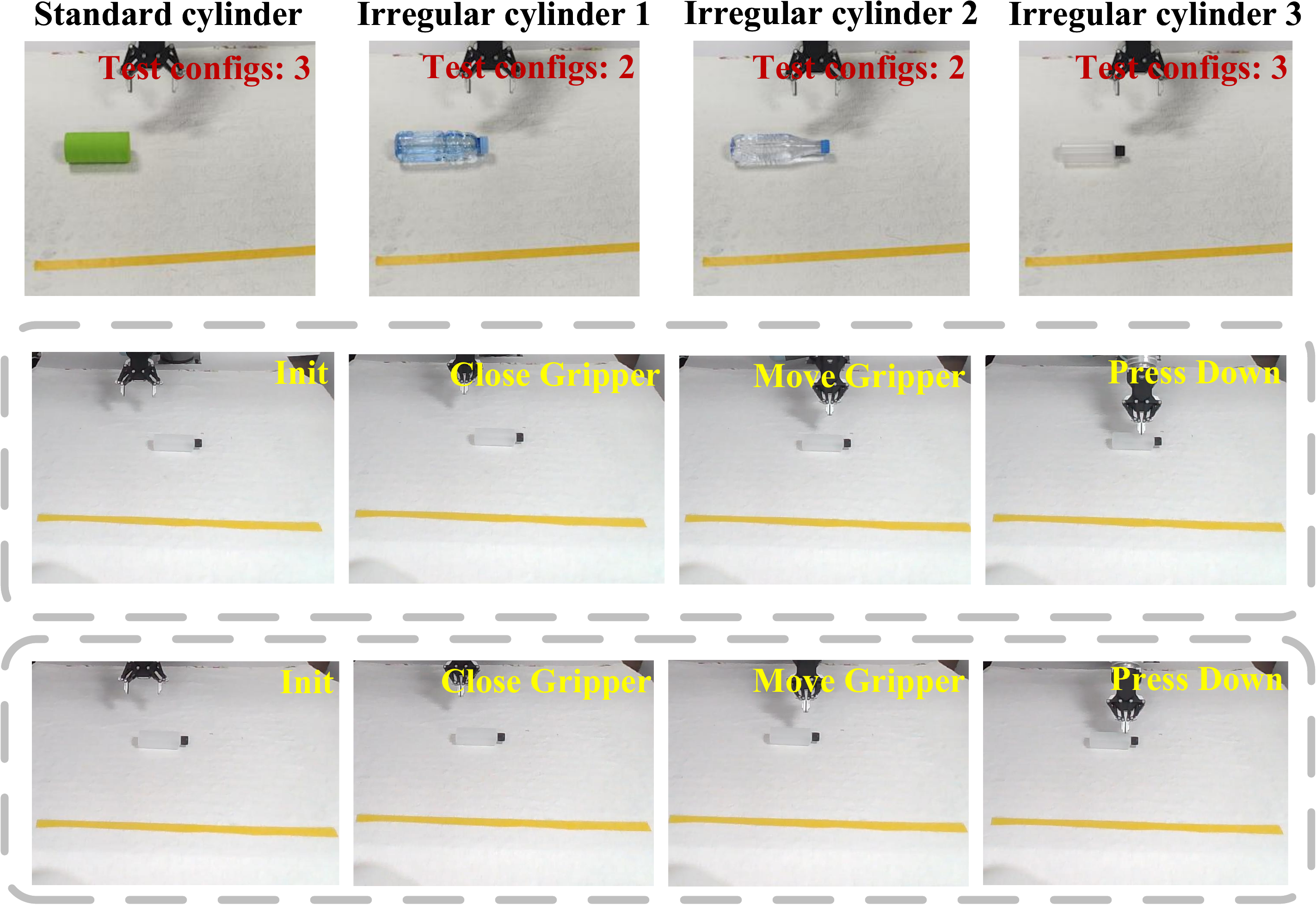}
	\caption{\textcolor{blue}{\textbf{Object Variation within the Base Task Sweep Cylinder.} The top row shows the training cylinder and three variants. Lower rows show selected DP3 press-like interactions. Success requires the object's leading edge to cross the yellow line. Examples illustrate behavior, not frequency. ``Test configs'' denotes trials per object type.}}
	\label{fig:sweep}
	\par\vspace{-0.25cm}
\end{figure}

\begin{figure}[!t]
	\centering
	\includegraphics[width=\columnwidth]{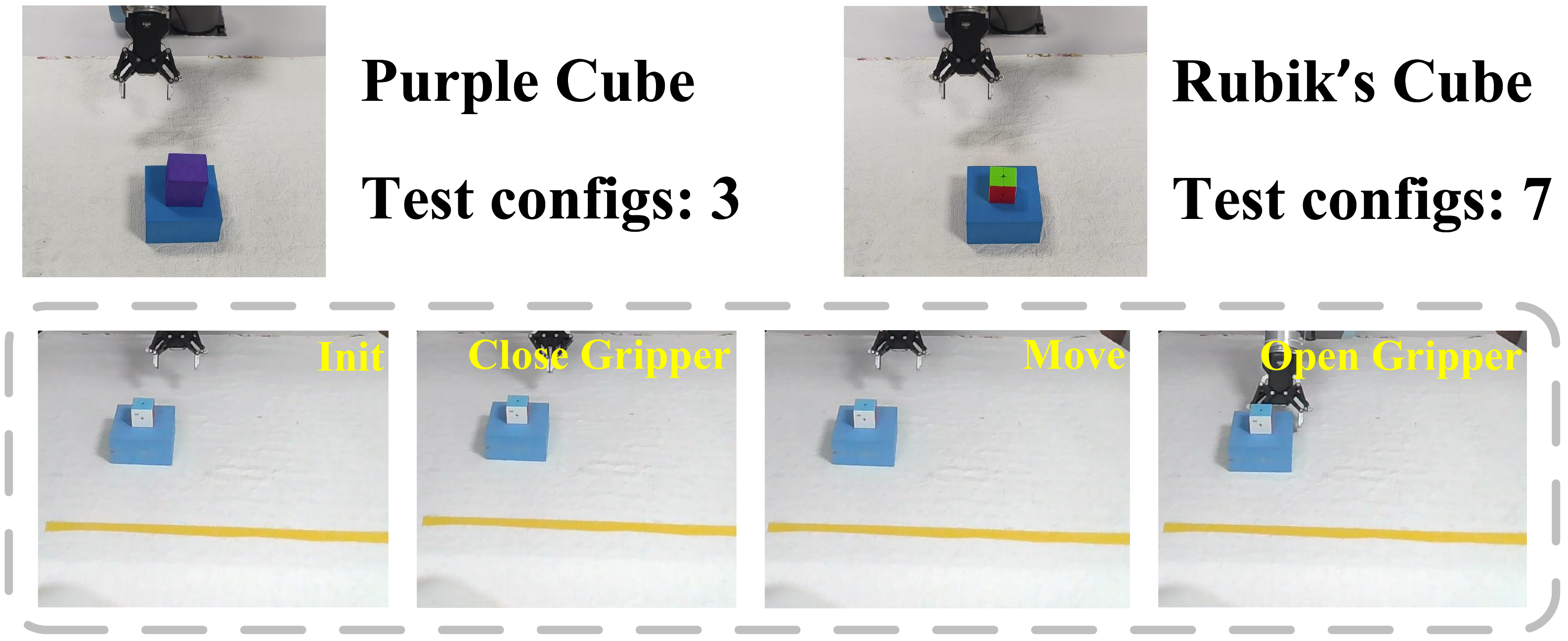}
	\caption{\textcolor{blue}{\textbf{Object Variation within Push Cube after 500-Step Adaptation.} Adaptation uses the purple cube, and testing also includes an unseen $2\times2$ Rubik's cube. The lower row shows one DP3 execution. Success requires pushing the small cube off its support. Examples illustrate behavior, not frequency. ``Test configs'' denotes trials per object type.}}
	\label{fig:push_cube}
	\vspace{-0.2cm}
\end{figure}

\textcolor{blue}{\textbf{Object Variation.} \Cref{fig:sweep,fig:push_cube} illustrate generalization to object instances absent from training or adaptation demonstrations. \textit{Sweep Cylinder} replaces the training cylinder with beverage and cosmetic bottles. \textit{Push Cube} additionally tests an unseen $2\times2$ Rubik's cube after adaptation using only the purple cube.}

\textcolor{blue}{\textbf{Task-Confusion Analysis.} We annotate all 70 rollouts per method, counting each rollout with task-inconsistent behavior at most once. A rollout is flagged if the gripper remains away from the target during the required interaction phase. The other criteria are stationary or cyclic actions without task progress, or an initial gripper state incompatible with the required interaction. DP3 has 2/10 flagged rollouts on both \textit{Press Button} and \textit{Sweep Cylinder}, and 1/10 on \textit{Push Cube}. For MAML-DP3, counts are 3/10 on \textit{Press Button}, 1/10 each on \textit{Sweep Cylinder} and \textit{Push Cube}, and 4/10 on \textit{Stack Cube}. All remaining task--method combinations have zero flagged rollouts. Overall rates are 7.14\% (5/70) for DP3, 12.86\% (9/70) for MAML-DP3, and 0\% (0/70) for DAMI.}

\textcolor{blue}{\textbf{DAMI Failure Cases.} The two failed trials involve grasp acquisition and placement completion. In \textit{Object Place}, the gripper closes without securing the fruit, which remains on the table. In \textit{Stack Cube}, grasping and lifting succeed, but the robot carries the cube past the support without placing it. The cube remains held beside the support at the end of the trial.}

\textcolor{blue}{\textbf{Illumination Changes.} We reduce illuminance from approximately 300 lx to 27 lx and test the same trained DAMI policy without fine-tuning, using ten trials per base task. Illuminance is estimated using the Coolexp smartphone light-meter application (version 2.1), with the same phone, measurement position, and orientation. Mean success falls from 98\% (49/50 trials) to 92\% (46/50). Only \textit{Sweep Cylinder} changes, from 100\% to 70\%, indicating task-dependent sensitivity to the tested illumination reduction.}

\textcolor{blue}{\textbf{Camera-View Changes.} We evaluate a DAMI policy trained using the original camera view, with current observations and reference demonstrations both obtained from the changed test view. The overall three-dimensional camera rotation is approximately $13^\circ$, and the optical-axis angle to the work surface increases from approximately $74^\circ$ to $76^\circ$. Each of the five base tasks uses ten trials. Success is 80\% on \textit{Sweep Cylinder}, 20\% on \textit{Close Drawer}, 10\% on \textit{Push Ball}, and 0\% on \textit{Object Place} and \textit{Press Button}. The mean is 22\% (11/50 trials), identifying the changed viewing conditions as a remaining generalization challenge.}

\subsection{Ablation Study}

\subsubsection{\textcolor{blue}{Controlled Component Configurations}}
\textcolor{blue}{We evaluate seven configurations under the within-update episode-disjoint meta-training protocol in Section IV-A, using the same evaluation seed. All retain MAML training, current observations, language, and complete reference demonstrations. Without VMT, a per-frame token MLP replaces its temporal Transformer and positional embedding. Without U2T, pooling and a $1536\!\rightarrow\!2048\!\rightarrow\!512$ MLP replace attention fusion. Without TCFM, additive conditioning, $h+\operatorname{Linear}(c)$, replaces task modulation at the first downsampling block, retaining other observation--timestep conditions. Single-module variants use both corresponding replacements, and names indicate the proposed modules retained.}

\begin{table}[!t]
\centering
\caption{\textcolor{blue}{ML10 Component Configurations. Task-Average Success Rates (\%) Use the Same Evaluation Seed.}}
\label{tab:ml10_component_configurations}
\begingroup\color{blue}\scriptsize
\setlength{\tabcolsep}{2pt}\renewcommand{\arraystretch}{1.12}
\begin{tabular*}{\columnwidth}{@{\extracolsep{\fill}}cccccc@{}}
\toprule
Configuration & VMT & U2T & TCFM & Base & Novel \\
\midrule
VMT only & $\checkmark$ & -- & -- & $87.60$ & $62.00$ \\
U2T only & -- & $\checkmark$ & -- & $89.00$ & $60.40$ \\
TCFM only & -- & -- & $\checkmark$ & $89.80$ & $59.40$ \\
VMT + U2T & $\checkmark$ & $\checkmark$ & -- & $89.00$ & $59.20$ \\
VMT + TCFM & $\checkmark$ & -- & $\checkmark$ & $90.00$ & $61.20$ \\
U2T + TCFM & -- & $\checkmark$ & $\checkmark$ & $88.10$ & $59.60$ \\
Full DAMI & $\checkmark$ & $\checkmark$ & $\checkmark$ & $\mathbf{90.70}$ & $\mathbf{62.40}$ \\
\bottomrule
\end{tabular*}
\par\vspace{-0.1cm}
\endgroup
\end{table}

\textcolor{blue}{Full DAMI has the highest base- and novel-task means in \Cref{tab:ml10_component_configurations}, at 90.70\% and 62.40\%, respectively. With the other two modules retained, removing VMT, U2T, or TCFM lowers novel-task means by 2.80, 1.20, or 3.20 percentage points, respectively. These differences support each module's contribution within the complete architecture. However, adding TCFM to VMT-only or U2T-only lowers the mean by 0.80 points, showing that its benefit depends on the module combination. The VMT-only configuration differs from the VMT variant using simple fusion in \Cref{tab:ml10_matched_conditioning}, which uses global concatenation and standard DP3 FiLM.}

\subsubsection{\textcolor{blue}{Training, Language, and VMT Ablations}}
\textcolor{blue}{\Cref{tab:ml10_vmt_targeted} examines training, language, and VMT representations using means and sample SDs across three evaluation seeds. Without MAML, ordinary diffusion BC replaces meta-training, retaining the full network, inputs, and 200 novel-task adaptation updates. Without language, text features are zeroed during training, adaptation, and inference. All VMT controls retain U2T, TCFM, language, current observations, and supervised action targets. Visual-only demonstrations retain point clouds but zero reference states and actions, whereas action-only demonstrations retain expert actions but zero reference-observation features. Action shuffling preserves observation order but permutes reference actions, disrupting action timing and observation--action correspondence. Permutations are resampled at each training encoding and fixed by rollout seed during evaluation. Removing VMT positional embeddings retains input order, the VMT Transformer, and U2T positional parameters.}

\begin{table}[!t]
\centering
\caption{\textcolor{blue}{ML10 Training, Language, and VMT Ablations. Success Rates (\%) Are Means $\pm$ Sample SDs across Three Evaluation Seeds.}}
\label{tab:ml10_vmt_targeted}
\begingroup\color{blue}\scriptsize
\setlength{\tabcolsep}{2pt}\renewcommand{\arraystretch}{1.12}
\begin{tabular*}{\columnwidth}{@{\extracolsep{\fill}}ccc@{}}
\toprule
Configuration & Base & Novel \\
\midrule
Full DAMI & $87.73 \pm 2.95$ & $\mathbf{57.93 \pm 4.31}$ \\
Without MAML training & $86.40 \pm 2.91$ & $55.47 \pm 2.32$ \\
Without language input & $87.10 \pm 4.50$ & $54.27 \pm 3.23$ \\
\midrule
Visual-only demonstration & $\mathbf{87.87 \pm 2.75}$ & $56.20 \pm 3.30$ \\
Action-only demonstration & $87.70 \pm 2.98$ & $56.73 \pm 4.12$ \\
Shuffled demonstration actions & $87.50 \pm 4.09$ & $55.40 \pm 6.54$ \\
\makecell[c]{Without VMT\\positional embedding} & $86.73 \pm 1.80$ & $56.60 \pm 4.28$ \\
\bottomrule
\end{tabular*}
\par\vspace{-0.25cm}
\endgroup
\end{table}

\textcolor{blue}{Full DAMI has the highest novel-task mean in \Cref{tab:ml10_vmt_targeted}, at 57.93\%. Without MAML or language, means fall to 55.47\% and 54.27\%, respectively, supporting both components for novel-task adaptation. Shuffling reference actions and removing VMT positional embeddings lower means to 55.40\% and 56.60\%, respectively. These results support retaining observation--action correspondence and temporal encoding in VMT. The visual-only variant has a slightly higher base-task mean.}

\subsection{\textcolor{blue}{Adaptation Budgets and Computational Efficiency}}
\subsubsection{\textcolor{blue}{Effect of Fine-Tuning Budget}}
\textcolor{blue}{On the five ML10 novel tasks, we vary adaptation updates for full DAMI trained under the original sampling protocol. We reuse the same pretrained initializations, demonstrations, and evaluation conditions without additional meta-training.}

\begin{table}[!t]
\centering
\caption{\textcolor{blue}{ML10 Novel-Task Success Rates (\%) across Fine-Tuning Budgets (Top-5 Mean $\pm$ Sample SD).}}
\label{tab:ml10_adaptation_budget}
\begingroup\color{blue}\footnotesize
\setlength{\tabcolsep}{5pt}
\begin{tabular}{cc}
\toprule
Updates & Novel tasks \\
\midrule
0 & $7.80 \pm 1.39$ \\
1 & $14.40 \pm 2.20$ \\
5 & $23.20 \pm 1.04$ \\
20 & $23.20 \pm 0.53$ \\
50 & $35.40 \pm 5.19$ \\
100 & $54.73 \pm 4.10$ \\
300 & $56.27 \pm 2.05$ \\
500 & $56.20 \pm 4.20$ \\
\midrule
200 (original evaluation) & $56.80 \pm 2.40$ \\
\bottomrule
\end{tabular}
\endgroup
\vspace{-0.4cm}
\end{table}

\textcolor{blue}{In \Cref{tab:ml10_adaptation_budget}, novel-task mean success rises from 7.80\% without updates to 23.20\% at five updates and 54.73\% at 100 updates. The original 200-step reference reaches 56.80\%, compared with 56.27\% at 300 updates and 56.20\% at 500. These top-5 results support retaining 200 updates for the reported simulation setting.}

\textcolor{blue}{For real-world \textit{Push Cube} and \textit{Stack Cube}, we test 100, 300, and 800 updates with ten trials per task and budget (\Cref{tab:real_adaptation_budget}). The main 500-step reference from \Cref{tab:real-task-success} gives 100\% and 90\%, respectively. At the additional budgets, \textit{Push Cube} ranges from 70\% to 100\%, whereas \textit{Stack Cube} reaches only 10--20\%. These results support the reported 500-step budget for the tested tasks and show that additional updates do not consistently improve success.}

\begin{table}[!t]
\centering
\caption{\textcolor{blue}{Real-World DAMI Success Rates (\%) across Fine-Tuning Budgets. Parentheses Give Successes/Trials.}}
\label{tab:real_adaptation_budget}
\begingroup\color{blue}\footnotesize
\setlength{\tabcolsep}{4pt}\renewcommand{\arraystretch}{1.1}
\begin{tabular}{ccc}
\toprule
Updates & Push Cube & Stack Cube \\
\midrule
100 & 100 (10/10) & 10 (1/10) \\
300 & 90 (9/10) & 10 (1/10) \\
800 & 70 (7/10) & 20 (2/10) \\
\midrule
500 (main evaluation) & 100 (10/10) & 90 (9/10) \\
\bottomrule
\end{tabular}
\par\vspace{-0.25cm}
\endgroup
\end{table}

\subsubsection{\textcolor{blue}{Effect of Demonstration Budget}}
\textcolor{blue}{For pretrained ML10 DAMI, we randomly select nested one- and five-demonstration subsets per task and seed. Base tasks change only the reference candidate pool, retaining pretraining data and weights. Novel tasks use 200 updates, drawing references and support windows from the same subset. Normalization inherits checkpoint statistics without refitting, with affine parameters following the original optimizer settings.}

\textcolor{blue}{\Cref{tab:ml10_demonstration_budget} uses three seeds controlling subset selection, adaptation, and evaluation. Base-task means remain similar, whereas novel-task means rise from 45.20\% with one demonstration to 54.20\% with five. The latter is 2.60 points below the original ten-demonstration reference of 56.80\%, retaining much of its performance with fewer trajectories.}

\begin{table}[!t]
\centering
\caption{\textcolor{blue}{ML10 Success Rates (\%) across Demonstration Budgets (Top-5 Mean $\pm$ Sample SD).}}
\label{tab:ml10_demonstration_budget}
\begingroup\color{blue}\footnotesize
\setlength{\tabcolsep}{1.5pt}\renewcommand{\arraystretch}{1.08}
\begin{tabular*}{\columnwidth}{@{\extracolsep{\fill}}cccc}
\toprule
\multicolumn{4}{c}{\textit{Base tasks, direct inference}} \\
\midrule
Task & 1 demo & 5 demos & 10 demos (ref.) \\
\midrule
Door Open & $90.33 \pm 11.72$ & $90.33 \pm 10.79$ & $92.67 \pm 2.08$ \\
\makecell[c]{Button Press\\Topdown} & $100.00 \pm 0.00$ & $100.00 \pm 0.00$ & $100.00 \pm 0.00$ \\
Drawer Close & $100.00 \pm 0.00$ & $100.00 \pm 0.00$ & $100.00 \pm 0.00$ \\
Reach & $86.00 \pm 3.46$ & $85.67 \pm 2.89$ & $58.67 \pm 5.03$ \\
Window Open & $99.67 \pm 0.58$ & $99.67 \pm 0.58$ & $100.00 \pm 0.00$ \\
Basketball & $100.00 \pm 0.00$ & $100.00 \pm 0.00$ & $100.00 \pm 0.00$ \\
Peg Insert Side & $74.67 \pm 9.71$ & $75.00 \pm 12.49$ & $88.00 \pm 8.54$ \\
Sweep & $84.00 \pm 10.15$ & $84.00 \pm 10.15$ & $87.33 \pm 3.51$ \\
Pick Place & $66.00 \pm 8.19$ & $66.67 \pm 9.71$ & $70.33 \pm 2.31$ \\
Push & $68.33 \pm 3.79$ & $68.33 \pm 4.73$ & $75.33 \pm 12.34$ \\
\midrule
\textbf{Average} & $86.90 \pm 2.95$ & $86.97 \pm 2.94$ & $87.23 \pm 2.02$ \\
\bottomrule
\end{tabular*}
\par\smallskip
\begin{tabular*}{\columnwidth}{@{\extracolsep{\fill}}cccc}
\toprule
\multicolumn{4}{c}{\textit{Novel tasks, 200-step adaptation}} \\
\midrule
Task & 1 demo & 5 demos & 10 demos (ref.) \\
\midrule
Door Close & $100.00 \pm 0.00$ & $100.00 \pm 0.00$ & $100.00 \pm 0.00$ \\
Drawer Open & $95.33 \pm 2.89$ & $99.00 \pm 1.73$ & $100.00 \pm 0.00$ \\
Lever Pull & $15.33 \pm 14.98$ & $39.33 \pm 11.24$ & $45.33 \pm 18.45$ \\
Sweep Into & $12.67 \pm 11.02$ & $24.33 \pm 3.06$ & $28.00 \pm 14.73$ \\
Shelf Place & $2.67 \pm 2.08$ & $8.33 \pm 6.03$ & $10.67 \pm 1.15$ \\
\midrule
\textbf{Average} & $45.20 \pm 5.10$ & $54.20 \pm 3.62$ & $56.80 \pm 2.40$ \\
\bottomrule
\end{tabular*}
\endgroup
\end{table}

\subsubsection{\textcolor{blue}{Wall-Clock Adaptation Time}}
\textcolor{blue}{We report mean fine-tuning times from one run per task across 45 tasks, excluding demonstration collection and subsequent policy execution. At 200 updates, DAMI takes 39.27 seconds versus 32.18 seconds for DP3, an additional 7.09 seconds (22.03\%). At 500 updates, respective times are 74.73 and 60.56 seconds, a difference of 14.17 seconds (23.40\%).}

\subsection{\textcolor{blue}{Reference-Demonstration Selection Sensitivity}}
\textcolor{blue}{For original-protocol ML10 and RLBench models, we vary inference-time references with fixed policy weights, using direct inference on base tasks and 200-step-adapted weights on novel tasks. Ten references share twenty initial conditions across three evaluation seeds, totaling 27,000 rollouts. \Cref{tab:reference_selection_sensitivity} summarizes seed-level means over references and tasks. Reference-selection variability is the sample SD across ten reference-specific success rates per task and seed, averaged over tasks and seeds.}

\begin{table}[!t]
\centering
\caption{\textcolor{blue}{Reference-Selection Sensitivity. Success Rates (\%) Are Means $\pm$ Sample SDs across Three Evaluation Seeds.}}
\label{tab:reference_selection_sensitivity}
\begingroup\color{blue}\footnotesize
\setlength{\tabcolsep}{2.5pt}\renewcommand{\arraystretch}{1.12}
\begin{tabular*}{\columnwidth}{@{\extracolsep{\fill}}cccc}
\toprule
Benchmark & \makecell[c]{Task\\group} & \makecell[c]{Success rate\\(\%)} & \makecell[c]{Across-reference\\SD (pp)} \\
\midrule
ML10 & Base & $82.72 \pm 0.64$ & $0.77$ \\
ML10 & Novel & $51.07 \pm 6.67$ & $1.34$ \\
RLBench & Base & $1.16 \pm 0.54$ & $0.30$ \\
RLBench & Novel & $10.93 \pm 3.50$ & $1.07$ \\
\bottomrule
\end{tabular*}
\endgroup
\end{table}

\textcolor{blue}{Average reference-selection SDs are below 1.5 percentage points in all four task groups, although individual tasks vary more. For example, \textit{Lever Pull} success ranges from 15\% to 35\% across references within one evaluation seed. The small RLBench base-task SD accompanies a low mean success of 1.16\%.}

\section{Conclusion}
\textcolor{blue}{We present DAMI, combining demonstration-conditioned 3D diffusion with full second-order MAML for base-task inference and novel-task adaptation. VMT encodes complete visual--action demonstrations, while U2T and TCFM fuse conditions and modulate policy features. Experiments span Meta-World, RLBench, and real robots, including real-world object generalization. Full DAMI achieves the highest novel-task top-5 mean among seven episode-disjoint ML10 component configurations. With VMT and U2T, TCFM improves over additive conditioning.}

\ifanonymous
\else
  \section*{Acknowledgments}
  This work is supported in part by the National Natural Science Foundation of China under Grant 62303447, Grant U23A20343, T2596045, in part by the Fundamental research project of SIA, No. 2023JC1K01 and No. 2023JC3K05 and in part by the Natural Science Foundation of Liaoning Province, No. 2024-MSBA-81.
\fi



\makeatletter
\newcommand{\DAMI@referencecolor}{%
  \par
  \color{black}%
}
\let\DAMI@originalbibitem\@bibitem
\def\@bibitem#1{%
  \DAMI@referencecolor%
  \DAMI@originalbibitem{#1}%
}
\let\DAMI@originallbibitem\@lbibitem
\def\@lbibitem[#1]#2{%
  \DAMI@referencecolor%
  \DAMI@originallbibitem[#1]{#2}%
}
\makeatother

\bibliographystyle{IEEEtran}
\bibliography{IEEEabrv,IEEEexample}

@IEEEtranBSTCTL{IEEEexample:BSTcontrol,
  CTLuse_article_number     = "yes",
  CTLuse_paper              = "yes",
  CTLuse_forced_etal        = "no",
  CTLmax_names_forced_etal  = "10",
  CTLnames_show_etal        = "1",
  CTLuse_alt_spacing        = "yes",
  CTLalt_stretch_factor     = "4",
  CTLdash_repeated_names    = "no",
  CTLname_format_string     = "{f.~}{vv~}{ll}{, jj}",
  CTLname_latex_cmd         = ""
}

@inproceedings{ji2025robobrain,
	title={Robobrain: A unified brain model for robotic manipulation from abstract to concrete},
	author={Ji, Yuheng and Tan, Huajie and Shi, Jiayu and Hao, Xiaoshuai and Zhang, Yuan and Zhang, Hengyuan and Wang, Pengwei and Zhao, Mengdi and Mu, Yao and An, Pengju and others},
	booktitle={Proceedings of the Computer Vision and Pattern Recognition Conference},
	pages={1724--1734},
	year={2025}
}

@article{li2025robotic,
	title={Robotic manipulation via imitation learning: Taxonomy, evolution, benchmark, and challenges},
	author={Li, Zezeng and Chapin, Alexandre and Xiang, Enda and Yang, Rui and Machado, Bruno and Lei, Na and Dellandrea, Emmanuel and Huang, Di and Chen, Liming},
	journal={arXiv preprint arXiv:2508.17449},
	year={2025},
	doi={10.48550/arXiv.2508.17449},
	url={https://arxiv.org/abs/2508.17449}
}

@inproceedings{jiang2025dexmimicgen,
	title={Dexmimicgen: Automated data generation for bimanual dexterous manipulation via imitation learning},
	author={Jiang, Zhenyu and Xie, Yuqi and Lin, Kevin and Xu, Zhenjia and Wan, Weikang and Mandlekar, Ajay and Fan, Linxi Jim and Zhu, Yuke},
	booktitle={2025 IEEE International Conference on Robotics and Automation (ICRA)},
	pages={16923--16930},
	year={2025},
	organization={IEEE}
}

@inproceedings{goyal2023rvt,
	title={Rvt: Robotic view transformer for 3d object manipulation},
	author={Goyal, Ankit and Xu, Jie and Guo, Yijie and Blukis, Valts and Chao, Yu-Wei and Fox, Dieter},
	booktitle={Conference on Robot Learning},
	pages={694--710},
	year={2023},
	organization={PMLR}
}

@inproceedings{gervet2023act3d,
	title={Act3D: 3D Feature Field Transformers for Multi-Task Robotic Manipulation},
	author={Gervet, Theophile and Xian, Zhou and Gkanatsios, Nikolaos and Fragkiadaki, Katerina},
	booktitle={Proceedings of The 7th Conference on Robot Learning},
	pages={3949--3965},
	year={2023},
	organization={PMLR}
}

@inproceedings{jang2025dreamgen,
	title={DreamGen: Unlocking Generalization in Robot Learning through Video World Models},
	author={Jang, Joel and Ye, Seonghyeon and Lin, Zongyu and Xiang, Jiannan and Bjorck, Johan and Fang, Yu and Hu, Fengyuan and Huang, Spencer and Kundalia, Kaushil and Lin, Yen-Chen and others},
	booktitle={Proceedings of The 9th Conference on Robot Learning},
	pages={5170--5194},
	year={2025},
	organization={PMLR}
}

@inproceedings{chen2025vividex,
	title={Vividex: Learning vision-based dexterous manipulation from human videos},
	author={Chen, Zerui and Chen, Shizhe and Arlaud, Etienne and Laptev, Ivan and Schmid, Cordelia},
	booktitle={2025 IEEE International Conference on Robotics and Automation (ICRA)},
	pages={3336--3343},
	year={2025},
	organization={IEEE}
}

@inproceedings{liu2025one,
	title={One-shot manipulation strategy learning by making contact analogies},
	author={Liu, Yuyao and Mao, Jiayuan and Tenenbaum, Joshua B and Lozano-P{\'e}rez, Tom{\'a}s and Kaelbling, Leslie Pack},
	booktitle={2025 IEEE International Conference on Robotics and Automation (ICRA)},
	pages={15387--15393},
	year={2025},
	organization={IEEE}
}

@inproceedings{kedia2025one,
	title={One-shot imitation under mismatched execution},
	author={Kedia, Kushal and Dan, Prithwish and Chao, Angela and Pace, Maximus A and Choudhury, Sanjiban},
	booktitle={2025 IEEE International Conference on Robotics and Automation (ICRA)},
	pages={15649--15656},
	year={2025},
	organization={IEEE}
}

@inproceedings{icilgd_iclr25,
	title = {Instant Policy: In-Context Imitation Learning via Graph Diffusion},
	author = {Vitalis Vosylius and Edward Johns},
	booktitle = {The Thirteenth International Conference on Learning Representations (ICLR)},
	year = {2025}
}

@inproceedings{oh2025robust,
	title={Robust Instant Policy: Leveraging Student's t-Regression Model for Robust In-context Imitation Learning of Robot Manipulation},
	author={Oh, Hanbit and Salcedo-V{\'a}zquez, Andrea M and Ramirez-Alpizar, Ixchel G and Domae, Yukiyasu},
	booktitle={2025 IEEE/RSJ International Conference on Intelligent Robots and Systems (IROS)},
	pages={7973--7980},
	year={2025},
	doi={10.1109/IROS60139.2025.11247005}
}

@inproceedings{hou20254d,
	title={4D Visual Pre-training for Robot Learning},
	author={Hou, Chengkai and Ze, Yanjie and Fu, Yankai and Gao, Zeyu and Hu, Songbo and Yu, Yue and Zhang, Shanghang and Xu, Huazhe},
	booktitle={Proceedings of the IEEE/CVF International Conference on Computer Vision},
	pages={8451--8461},
	year={2025}
}

@inproceedings{Ze2024DP3,
	title={3D Diffusion Policy: Generalizable Visuomotor Policy Learning via Simple 3D Representations},
	author={Yanjie Ze and Gu Zhang and Kangning Zhang and Chenyuan Hu and Muhan Wang and Huazhe Xu},
	booktitle={Proceedings of Robotics: Science and Systems (RSS)},
	year={2024}
}

@inproceedings{cao2025mamba,
  title={Mamba policy: Towards efficient 3d diffusion policy with hybrid selective state models},
  author={Cao, Jiahang and Zhang, Qiang and Sun, Jingkai and Wang, Jiaxu and Cheng, Hao and Li, Yulin and Ma, Jun and Wu, Kun and Xu, Zhiyuan and Shao, Yecheng and others},
  booktitle={2025 IEEE/RSJ International Conference on Intelligent Robots and Systems (IROS)},
  pages={11359--11366},
  year={2025},
  organization={IEEE}
}

@inproceedings{liu2025spatial,
	title={Spatial-Temporal Aware Visuomotor Diffusion Policy Learning},
	author={Liu, Zhenyang and Wang, Yikai and Wang, Kuanning and Liang, Longfei and Xue, Xiangyang and Fu, Yanwei},
	booktitle={Proceedings of the IEEE/CVF International Conference on Computer Vision},
	pages={7122--7131},
	year={2025}
}

@inproceedings{finn2017model,
	title={Model-agnostic meta-learning for fast adaptation of deep networks},
	author={Finn, Chelsea and Abbeel, Pieter and Levine, Sergey},
	booktitle={International conference on machine learning},
	pages={1126--1135},
	year={2017},
	organization={PMLR}
}

@inproceedings{fang2025airexo,
	title={AirExo-2: Scaling up Generalizable Robotic Imitation Learning with Low-Cost Exoskeletons},
	author={Fang, Hongjie and Wang, Chenxi and Wang, Yiming and Chen, Jingjing and Xia, Shangning and Lv, Jun and He, Zihao and Yi, Xiyan and Guo, Yunhan and Zhan, Xinyu and others},
	booktitle={Proceedings of The 9th Conference on Robot Learning},
	pages={198--220},
	year={2025},
	organization={PMLR}
}

@inproceedings{jang2022bc,
	title={Bc-z: Zero-shot task generalization with robotic imitation learning},
	author={Jang, Eric and Irpan, Alex and Khansari, Mohi and Kappler, Daniel and Ebert, Frederik and Lynch, Corey and Levine, Sergey and Finn, Chelsea},
	booktitle={Conference on Robot Learning},
	pages={991--1002},
	year={2022},
	organization={PMLR}
}

@article{kuang2024ram,
  title={Ram: Retrieval-based affordance transfer for generalizable zero-shot robotic manipulation},
  author={Kuang, Yuxuan and Ye, Junjie and Geng, Haoran and Mao, Jiageng and Deng, Congyue and Guibas, Leonidas and Wang, He and Wang, Yue},
  journal={arXiv preprint arXiv:2407.04689},
  year={2024}
}

@inproceedings{pan2025omnimanip,
	title={Omnimanip: Towards general robotic manipulation via object-centric interaction primitives as spatial constraints},
	author={Pan, Mingjie and Zhang, Jiyao and Wu, Tianshu and Zhao, Yinghao and Gao, Wenlong and Dong, Hao},
	booktitle={Proceedings of the Computer Vision and Pattern Recognition Conference},
	pages={17359--17369},
	year={2025}
}

@article{xu2025funcanon,
	title={{FUNCanon}: Learning Pose-Aware Action Primitives via Functional Object Canonicalization for Generalizable Robotic Manipulation},
	author={Xu, Hongli and Zhang, Lei and Hu, Xiaoyue and Zhong, Boyang and Bai, Kaixin and M{\'a}rton, Zolt{\'a}n-Csaba and Bing, Zhenshan and Chen, Zhaopeng and Knoll, Alois Christian and Zhang, Jianwei},
	journal={arXiv preprint arXiv:2509.19102},
	year={2025},
	doi={10.48550/arXiv.2509.19102},
	url={https://arxiv.org/abs/2509.19102}
}

@inproceedings{zhang2024imop,
	title={One-shot imitation learning with invariance matching for robotic manipulation},
	author={Zhang, Xinyu and Boularias, Abdeslam},
	booktitle={Proceedings of Robotics: Science and Systems},
	year={2024},
	doi={10.15607/RSS.2024.XX.134},
	url={https://roboticsconference.org/2024/program/papers/134/}
}

@inproceedings{goyal2024rvt,
	title={RVT-2: Learning Precise Manipulation from Few Demonstrations},
	author={Goyal, Ankit and Blukis, Valts and Xu, Jie and Guo, Yijie and Chao, Yu-Wei and Fox, Dieter},
	booktitle={Proceedings of Robotics: Science and Systems},
	year={2024}
}

@inproceedings{ze2023gnfactor,
	title={Gnfactor: Multi-task real robot learning with generalizable neural feature fields},
	author={Ze, Yanjie and Yan, Ge and Wu, Yueh-Hua and Macaluso, Annabella and Ge, Yuying and Ye, Jianglong and Hansen, Nicklas and Li, Li Erran and Wang, Xiaolong},
	booktitle={Conference on robot learning},
	pages={284--301},
	year={2023},
	organization={PMLR}
}

@inproceedings{lu2024manigaussian,
	title={Manigaussian: Dynamic gaussian splatting for multi-task robotic manipulation},
	author={Lu, Guanxing and Zhang, Shiyi and Wang, Ziwei and Liu, Changliu and Lu, Jiwen and Tang, Yansong},
	booktitle={European Conference on Computer Vision},
	pages={349--366},
	year={2024},
	organization={Springer}
}

@article{ke20243d,
  title={3d diffuser actor: Policy diffusion with 3d scene representations},
  author={Ke, Tsung-Wei and Gkanatsios, Nikolaos and Fragkiadaki, Katerina},
  journal={arXiv preprint arXiv:2402.10885},
  year={2024}
}

@inproceedings{tian2025pdfactor,
	title={PDFactor: Learning Tri-Perspective View Policy Diffusion Field for Multi-Task Robotic Manipulation},
	author={Tian, Jingyi and Wang, Le and Zhou, Sanping and Wang, Sen and Li, Jiayi and Sun, Haowen and Tang, Wei},
	booktitle={Proceedings of the Computer Vision and Pattern Recognition Conference},
	pages={15757--15767},
	year={2025}
}

@inproceedings{wang2025flowram,
	title={FlowRAM: Grounding Flow Matching Policy with Region-Aware Mamba Framework for Robotic Manipulation},
	author={Wang, Sen and Wang, Le and Zhou, Sanping and Tian, Jingyi and Li, Jiayi and Sun, Haowen and Tang, Wei},
	booktitle={Proceedings of the Computer Vision and Pattern Recognition Conference},
	pages={12176--12186},
	year={2025}
}

@inproceedings{radford2021learning,
	title={Learning transferable visual models from natural language supervision},
	author={Radford, Alec and Kim, Jong Wook and Hallacy, Chris and Ramesh, Aditya and Goh, Gabriel and Agarwal, Sandhini and Sastry, Girish and Askell, Amanda and Mishkin, Pamela and Clark, Jack and others},
	booktitle={International conference on machine learning},
	pages={8748--8763},
	year={2021},
	organization={PmLR}
}

@inproceedings{yu2020meta,
	title={Meta-world: A benchmark and evaluation for multi-task and meta reinforcement learning},
	author={Yu, Tianhe and Quillen, Deirdre and He, Zhanpeng and Julian, Ryan and Hausman, Karol and Finn, Chelsea and Levine, Sergey},
	booktitle={Conference on robot learning},
	pages={1094--1100},
	year={2020},
	organization={PMLR}
}

@article{james2020rlbench,
	title={{RLBench}: The Robot Learning Benchmark \& Learning Environment},
	author={James, Stephen and Ma, Zicong and Arrojo, David Rovick and Davison, Andrew J.},
	journal={IEEE Robotics and Automation Letters},
	volume={5},
	number={2},
	pages={3019--3026},
	year={2020},
	doi={10.1109/LRA.2020.2974707}
}

@inproceedings{xia2025cage,
	title={Cage: Causal attention enables data-efficient generalizable robotic manipulation},
	author={Xia, Shangning and Fang, Hongjie and Lu, Cewu and Fang, Hao-Shu},
	booktitle={2025 IEEE International Conference on Robotics and Automation (ICRA)},
	pages={13242--13249},
	year={2025},
	organization={IEEE}
}

@article{su2026freqpolicy,
  title={Freqpolicy: Efficient flow-based visuomotor policy via frequency consistency},
  author={Su, Yifei and Liu, Ning and Chen, Dong and Zhao, Zhen and Wu, Kun and Li, Meng and Xu, Zhiyuan and Che, Zhengping and Tang, Jian},
  journal={Advances in Neural Information Processing Systems},
  volume={38},
  pages={27769--27797},
  year={2026}
}

@inproceedings{zhang2025flowpolicy, 
	title={Flowpolicy: Enabling fast and robust 3d flow-based policy via consistency flow matching for robot manipulation},
	author={Zhang, Qinglun and Liu, Zhen and Fan, Haoqiang and Liu, Guanghui and Zeng, Bing and Liu, Shuaicheng},
	booktitle={Proceedings of the AAAI Conference on Artificial Intelligence},
	pages={14754--14762},
	year={2025}
}

@article{zhong2026freqpolicy,
  title={Freqpolicy: Frequency autoregressive visuomotor policy with continuous tokens},
  author={Zhong, Yiming and Liu, Yumeng and Xiao, Chuyang and Yang, Zemin and Wang, Youzhuo and Zhu, Yufei and Shi, Ye and Sun, Yujing and Zhu, Xinge and Ma, Yuexin},
  journal={Advances in Neural Information Processing Systems},
  volume={38},
  pages={56493--56526},
  year={2026}
}

@inproceedings{ronneberger2015u,
	title={U-net: Convolutional networks for biomedical image segmentation},
	author={Ronneberger, Olaf and Fischer, Philipp and Brox, Thomas},
	booktitle={International Conference on Medical image computing and computer-assisted intervention},
	pages={234--241},
	year={2015},
	organization={Springer}
}

@article{chi2025diffusion,
  title={Diffusion policy: Visuomotor policy learning via action diffusion},
  author={Chi, Cheng and Xu, Zhenjia and Feng, Siyuan and Cousineau, Eric and Du, Yilun and Burchfiel, Benjamin and Tedrake, Russ and Song, Shuran},
  journal={The International Journal of Robotics Research},
  volume={44},
  number={10-11},
  pages={1684--1704},
  year={2025},
  publisher={Sage Publications Sage UK: London, England}
}

@inproceedings{perez2018film,
	title={Film: Visual reasoning with a general conditioning layer},
	author={Perez, Ethan and Strub, Florian and De Vries, Harm and Dumoulin, Vincent and Courville, Aaron},
	booktitle={Proceedings of the AAAI conference on artificial intelligence},
	year={2018}
}

@article{qi2017pointnet++,
	title={Pointnet++: Deep hierarchical feature learning on point sets in a metric space},
	author={Qi, Charles Ruizhongtai and Yi, Li and Su, Hao and Guibas, Leonidas J},
	journal={Advances in neural information processing systems},
	volume={30},
	year={2017}
}

@article{kerbl20233d,
	title={3D Gaussian splatting for real-time radiance field rendering.},
	author={Kerbl, Bernhard and Kopanas, Georgios and Leimk{\"u}hler, Thomas and Drettakis, George},
	journal={ACM Trans. Graph.},
	volume={42},
	number={4},
	pages={139--1},
	year={2023}
}

@inproceedings{team2024octo,
	title={Octo: An Open-Source Generalist Robot Policy},
	author={{Octo Model Team} and Ghosh, Dibya and Walke, Homer and Pertsch, Karl and Black, Kevin and Mees, Oier and Dasari, Sudeep and Hejna, Joey and Kreiman, Tobias and Xu, Charles and Luo, Jianlan and Tan, {You Liang} and Chen, {Lawrence Yunliang} and Sanketi, Pannag R. and Vuong, Quan and Xiao, Ted and Sadigh, Dorsa and Finn, Chelsea and Levine, Sergey},
	booktitle={Proceedings of Robotics: Science and Systems},
	year={2024}
}

@inproceedings{wang2025one,
	title={One-shot video imitation via parameterized symbolic abstraction graphs},
	author={Wang, Jianren and Liu, Kangni and Guo, Dingkun and Xian, Zhou and Atkeson, Christopher G},
	booktitle={2025 IEEE International Conference on Robotics and Automation (ICRA)},
	pages={10552--10560},
	year={2025},
	organization={IEEE}
}

@inproceedings{ze2025generalizable,
	title={Generalizable humanoid manipulation with 3d diffusion policies},
	author={Ze, Yanjie and Chen, Zixuan and Wang, Wenhao and Chen, Tianyi and He, Xialin and Yuan, Ying and Peng, Xue Bin and Wu, Jiajun},
	booktitle={2025 IEEE/RSJ International Conference on Intelligent Robots and Systems (IROS)},
	pages={2873--2880},
	year={2025},
	organization={IEEE}
}

@article{lin2025adaptive,
  title={Adaptive Video-Conditioned Imitation Learning via Bidirectional Cross-Domain Skill Transfer},
  author={Lin, Zhenyang and Chen, Yurou and Li, Zhengwei and Zhang, Xianxiang and Liang, Bin and Liu, Zhiyong},
  journal={IEEE Transactions on Automation Science and Engineering},
  volume={22},
  pages={23214--23227},
  year={2025},
  publisher={IEEE}
}

@article{xu2025bikc+,
  title={BiKC+: Bimanual Hierarchical Imitation With Keypose-Conditioned Coordination-Aware Consistency Policies},
  author={Xu, Hang and Chen, Yizhou and Yu, Dongjie and Ren, Yi and Pan, Jia},
  journal={IEEE Transactions on Automation Science and Engineering},
  volume={23},
  pages={1064--1079},
  year={2025},
  publisher={IEEE}
}

@article{lin2026toward,
  title={Toward reliable imitation learning with limited expert demonstrations via search-based inverse dynamic learning},
  author={Lin, Zhiliang and Chen, Zhuangzhuang and Zhu, Guanming and Wang, Li and Li, Jianqiang},
  journal={IEEE Transactions on Automation Science and Engineering},
  year={2026},
  publisher={IEEE}
}

@article{wang2025robot,
  title={Robot Deformable Object Manipulation via NMPC-Generated Demonstrations in Deep Reinforcement Learning},
  author={Wang, Haoyuan and Dong, Zihao and Zhu, Tong and Lei, Hongliang and Shi, Weizhuang and Zhang, Zejia and Luo, Wei and Wan, Weiwei and Chen, Xinxing and Huang, Jian},
  journal={IEEE Transactions on Automation Science and Engineering},
  volume={22},
  pages={23566--23578},
  year={2025},
  publisher={IEEE}
}

@article{zhao2025robot,
  title={Robot Dexterous Grasping in Cluttered Scenes Based on Single-View Point Cloud},
  author={Zhao, Qingxing and Zheng, Minhua and Li, Zhaoxin and Huang, Shichang and Shi, Wen},
  journal={IEEE Transactions on Automation Science and Engineering},
  year={2025},
  publisher={IEEE}
}

@article{zhong2025region,
  title={Region-aware grasping for stacked workpieces: A 6D-wise label self-generation method and robust evaluation strategy},
  author={Zhong, Xungao and Gong, Tao and Yu, Junzhi and Luo, Jiaguo and Zhou, Chengxian and Zhong, Xunyu and Liu, Qiang},
  journal={IEEE Transactions on Automation Science and Engineering},
  volume={22},
  pages={16899--16912},
  year={2025},
  publisher={Institute of Electrical and Electronics Engineers (IEEE)}
}

@article{yu2025fast,
  title={Fast and accurate category-level object pose estimation without shape priors for robotic grasp detection},
  author={Yu, Sheng and Yin, Jian and Zhai, Di-Hua and Xia, Yuanqing},
  journal={IEEE Transactions on Automation Science and Engineering},
  year={2025},
  publisher={IEEE}
}

@article{song2025learning,
  title={Learning 6-dof fine-grained grasp detection based on part affordance grounding},
  author={Song, Yaoxian and Sun, Penglei and Jin, Piaopiao and Ren, Yi and Zheng, Yu and Li, Zhixu and Chu, Xiaowen and Zhang, Yue and Li, Tiefeng and Gu, Jason},
  journal={IEEE Transactions on Automation Science and Engineering},
  year={2025},
  publisher={IEEE}
}

@article{zhao2026affordance,
  title={Affordance-Guided Robotic Grasping via Multimodal Large Language Model Reasoning},
  author={Zhao, Zhou and Gao, Jie and Zheng, Dongyuan},
  journal={IEEE Transactions on Automation Science and Engineering},
  year={2026},
  publisher={IEEE}
}

@article{chen2025adp,
  title={ADP: Adaptive diffusion policy energizes robots thinking in both learning and practice},
  author={Chen, Dechao and Chen, Zhengwen and Zheng, Xiangyan and Xu, Weiling and Ma, Chencong and Mao, Chentao},
  journal={IEEE Transactions on Automation Science and Engineering},
  year={2025},
  publisher={IEEE}
}

@article{zhao2026robot,
  title={Robot Few-Shot Manipulation Skills Learning Based on Meta Imitation Learning and Mixture of Experts Model},
  author={Zhao, Jiahe and Liu, Jiahang and Su, Xiu and He, Bin and Jiang, Shuo},
  journal={IEEE Transactions on Automation Science and Engineering},
  year={2026},
  publisher={IEEE}
}

@article{luo2026cova,
  title={CoVA-IL: Zero-Shot Imitation Learning via Contrastive Viewpoint Alignment on Object-Centric Representation},
  author={Luo, Jiangtao and Yang, Shuo and Zheng, Chenchen and Fan, Jinqiu and Gao, Yang and Song, Ran and Zhang, Wei},
  journal={IEEE Transactions on Automation Science and Engineering},
  year={2026},
  publisher={IEEE}
}

@article{wu2025autolfd,
  title={Autolfd: Closing the loop for learning from demonstrations},
  author={Wu, Shaokang and Wang, Yijin and Huang, Yanlong},
  journal={IEEE Transactions on Automation Science and Engineering},
  volume={22},
  pages={11124--11138},
  year={2025},
  publisher={IEEE}
}

@article{liu2025rasp,
  title={RASP: Robot Active Scene Perception With Joint Viewpoint Planning and Depth Completion in Cluttered Environments},
  author={Liu, Yizhe and Jia, Tong and Zhang, Haiyu and Yang, Guowei and Wang, Hao and Chen, Dongyue},
  journal={IEEE Transactions on Automation Science and Engineering},
  year={2025},
  publisher={IEEE}
}

@article{ding2025dual,
  title={Dual graph attention networks for multi-view visual manipulation relationship detection and robotic grasping},
  author={Ding, Mengyuan and Liu, Yaxin and Shi, Yaorui and Lan, Xuguang and Zheng, Nanning},
  journal={IEEE Transactions on Automation Science and Engineering},
  year={2025},
  publisher={IEEE}
}

@inproceedings{fu2025icrt,
  title={{ICRT}: In-Context Imitation Learning via Next-Token Prediction},
  author={Fu, Max and Huang, Huang and Datta, Gaurav and Chen, Lawrence Yunliang and Panitch, William Chung-Ho and Liu, Fangchen and Li, Hui and Goldberg, Ken},
  booktitle={Proceedings of the IEEE International Conference on Robotics and Automation},
  year={2025},
  pages={5937--5944},
  url={https://icrt.dev/}
}

@inproceedings{li2025controlvla,
  title={{ControlVLA}: Few-Shot Object-Centric Adaptation for Pre-Trained Vision-Language-Action Models},
  author={Li, Puhao and Wu, Yingying and Xi, Ziheng and Li, Wanlin and Huang, Yuzhe and Zhang, Zhiyuan and Chen, Yinghan and Wang, Jianan and Zhu, Song-Chun and Liu, Tengyu and Huang, Siyuan},
  booktitle={Proceedings of the 9th Conference on Robot Learning},
  series={Proceedings of Machine Learning Research},
  volume={305},
  year={2025},
  pages={1898--1913},
  publisher={PMLR},
  url={https://proceedings.mlr.press/v305/li25c.html}
}

@inproceedings{liu2025rdt1b,
  title={{RDT-1B}: A Diffusion Foundation Model for Bimanual Manipulation},
  author={Liu, Songming and Wu, Lingxuan and Li, Bangguo and Tan, Hengkai and Chen, Huayu and Wang, Zhengyi and Xu, Ke and Su, Hang and Zhu, Jun},
  booktitle={Proceedings of the International Conference on Learning Representations},
  year={2025},
  url={https://arxiv.org/abs/2410.07864}
}

@inproceedings{park2026demodiffusion,
  title={{DemoDiffusion}: One-Shot Human Imitation Using Pre-Trained Diffusion Policy},
  author={Park, Sungjae and Bharadhwaj, Homanga and Tulsiani, Shubham},
  booktitle={Proceedings of the IEEE International Conference on Robotics and Automation},
  year={2026},
  url={https://demodiffusion.github.io/}
}

@inproceedings{finn2017visualmetaimitation,
  title={One-Shot Visual Imitation Learning via Meta-Learning},
  author={Finn, Chelsea and Yu, Tianhe and Zhang, Tianhao and Abbeel, Pieter and Levine, Sergey},
  booktitle={Proceedings of the 1st Annual Conference on Robot Learning},
  series={Proceedings of Machine Learning Research},
  volume={78},
  year={2017},
  pages={357--368},
  publisher={PMLR},
  url={https://proceedings.mlr.press/v78/finn17a.html}
}

@inproceedings{shah2023mutex,
  title={{MUTEX}: Learning Unified Policies from Multimodal Task Specifications},
  author={Shah, Rutav and Mart{\'i}n-Mart{\'i}n, Roberto and Zhu, Yuke},
  booktitle={Proceedings of the 7th Conference on Robot Learning},
  series={Proceedings of Machine Learning Research},
  volume={229},
  year={2023},
  pages={2663--2682},
  publisher={PMLR},
  url={https://proceedings.mlr.press/v229/shah23b.html}
}



\end{document}